\documentclass[preprint]{article}
\pdfoutput=1

\usepackage{natbib}
\usepackage{amsmath,amssymb,amsfonts}
\usepackage{algorithmic}
\usepackage{graphicx}
\usepackage{textcomp}
\usepackage{xcolor}
\usepackage{optprog}
\usepackage{mathtools}
\usepackage{placeins}
\usepackage{array,multirow,graphicx}
\usepackage{float}
\usepackage{multirow}
\usepackage{booktabs}
\usepackage{comment}
\usepackage[margin=2cm]{geometry}
\newtheorem{theorem}{Theorem}
\newtheorem{proof}{Proof}

\title{Ensemble pruning via an integer programming approach with diversity constraints}

\author{Marcelo Antônio Mendes Bastos$^1$ \and Humberto Brandão César de Oliveira$^2$ \and Cristiano Arbex Valle$^1$}

\date{}

\begin{document}

\maketitle

\begin{center} 
{\footnotesize

$^1$Departamento de Ci\^{e}ncia da Computa\c{c}\~{a}o, \\
 Universidade Federal de Minas Gerais, \\
 Belo Horizonte, MG, Brasil \\
 \{marcelo.bastos,arbex\}@dcc.ufmg.br \\ \vspace{0.3cm}
$^2$ Departamento de Ciência da Computação, \\
Universidade Federal de Alfenas, \\
Alfenas, MG, Brazil \\ humberto@bcc.unifal-mg.edu.br
}
\end{center}

\maketitle

\vspace{0.3cm}
\begin{abstract}
Ensemble learning combines multiple classifiers in the hope of obtaining better predictive performance. Empirical studies have shown that ensemble pruning, that is, choosing an appropriate subset of the available classifiers, can lead to comparable or better predictions than using all classifiers. In this paper, we consider a binary classification problem and propose an integer programming (IP) approach for selecting optimal classifier subsets. We propose a flexible objective function to adapt to desired criteria of different datasets. We also propose constraints to ensure minimum diversity levels in the ensemble. Despite the general case of IP being NP-Hard, state-of-the-art solvers are able to quickly obtain good solutions for datasets with up to 60000 data points. Our approach yields competitive results when compared to some of the best and most used pruning methods in literature. 

\end{abstract}


\section{Introduction}\label{intro}

Ensemble learning is a popular technique in the domain of machine learning. An ensemble is defined as the aggregation of multiple classifications into a single final decision. It is generally accepted in literature that the precision of an  ensemble tends to improve when compared to the behaviour of individual classifiers~\citep{zhang2006}. 

Well known aproaches for efficiently generating ensembles include Bagging (bootstrap aggregating) \citep{breiman1996} and Boosting \citep{freund1997}. The latter gave rise to popular variations such as Random Forests \citep{breiman2001} and extreme gradient boosting \citep{chen2016}. A general feature of these approaches is that all classifiers are considered in the aggregation. There are, however, theoretical and empirical studies which have shown that pruning an ensemble by selecting a subset of the classifiers can lead to comparable or better predictions \citep{lu2010,zhang2006}.

In this work we tackle the ensemble pruning problem by introducing an integer programming (IP) approach for choosing an optimal subset of binary classifiers. Our formulation optimises a weighted function of the patterns in the binary confusion matrix. This provides enough flexibility to ensure suitable optimisation criteria dependent on the properties of the underlying dataset. As our objective is performance-based we also introduce linear constraints that ensure minimum diversity levels in the ensemble. 

Despite several techniques for ensemble pruning having been previously proposed, we believe that our approach  contributes to the current knowledge in the field. IP modelling is a flexible tool, adaptable to particularities of different problems. One of the most important advantages in applying this tool to ensemble pruning is being able to combine performance and diversity criteria. Moreover an IP framework provides an exact method, as opposed to most algorithms in literature which are generally suboptimal.

The general IP problem is NP-Hard, however decades of research in algorithmic techniques have led to state-of-the-art solvers which are able to effectively solve several industrial-sized problems. In this paper we show that with such solvers we can find good solutions to relatively large problems in reasonable computational times. 

We compare our formulation to a full ensemble and six other well-known methods in literature. We report competitive results for publicly available datasets ranging from 195 to 60000 data points. 


The remainder of this paper is organised as follows. In Section \ref{sec:review} we give a brief overview of existing methods in ensemble learning. In Section \ref{sec:model} we present our optimisation approach and in Section \ref{sec:diversity} we amend it to enforce minimum diversity levels. Our computational experiments are shown in Section \ref{sec:results} and in Section \ref{sec:conclusion} we present our concluding remarks.

\section{Literature review}\label{sec:review}



The first step in an ensemble process is to generate a set of distinct classifiers that is hopefully precise and diverse. Highly correlated classifiers may hinder the potential benefit of using an ensemble.

Several techniques for ensuring diversity in classifiers have been proposed \citep{duin2002,cruz2018}. These include randomising classifiers, tuning parameters (e.g. pruning level in decision trees) or exploring different architectures (e.g. number of hidden layers in multi-layer perceptron neural networks). Combining distinct classifier models into heterogeneous ensembles is also a popular approach. Other diversification techniques include training classifiers with different distributions of the training set (e.g. the aforementioned Bagging and Boosting techniques) and with distinct subsets of features.

The next step is selecting an appropriate subset of the available classifiers. This selection can be static or dynamic. In dynamic selection, different subsets are chosen for different data points. The reasoning is that certain subsets may be more specialised in different parts of the feature space. For more details we refer the reader to \citep{britto2014,cruz2018}. Most works however apply static selection, where a single subset is chosen. Static selection policies can be based on ranking, clusters and optimisation.  

Ranking methods sort classifiers according to a fitness function. In general they greedily increase the subset size. In Kappa pruning \citep{margineantu1997}, every pair of classifiers is sorted according to a statistical measure of agreement. Pairs with the lowest agreement levels are selected. Reordering techniques \citep{martinez2006, martinez2007} in bagging classifiers have been used to build subensembles of increasing size. \cite{tsoumakas2004} rank heterogeneous classifiers according to a significance index. \cite{dai2016} proposed a dynamic programming approach to improve the computational efficiency of these methods.

Cluster methods first apply a clustering algorithm to separate classifiers according to some similarity measure and then prune each cluster separately to increase general diversity. Known clustering algorithms include k-means \citep{lazarevic2001,qiang2005}, where similarity is based on Euclidean distance, and hierarchical agglomerative clustering \citep{giacinto2000}, which employs probabilities. Different policies for choosing the number of clusters \citep{giacinto2000,lazarevic2001} have also been suggested.

Several optimisation methods for ensemble pruning have also been proposed, with most offering approximate solutions. The most popular method is hill climbing, which has been applied with several different fitness functions.  Some are based on performance \citep{fan2002,caruana2004} (e.g. accuracy), others on diversity \citep{margineantu1997, tang2006, partalas2008}. Two examples of diversity-based fitness functions are Complementariness \citep{martinez2004} and Concurrency \citep{banfield2005}. \cite{partalas2010} proposed the Uncertainty Weighted Accuracy, which as the name suggests takes into account uncertainty by weighting differently the importance of data points considered ``easy'' and ``hard'' to predict. \cite{partalas2006} employed a reinforcement learning-based approach for a greedy method based on diversity. \cite{zhang2006} and \cite{xu2012} proposed semi-definite programming approaches which consider trade-offs between accuracy and diversity. 


The last step in the procedure is combining classifiers into a single prediction, which is usually done through majority voting. For further details we refer the reader to \citep{kittler1998}.

\section{Formulation}
\label{sec:model}

Consider a binary classification problem where data points belong to classes 1 (positive) or 0 (negative). In this section we present an IP formulation for choosing an optimal subset of binary classifiers.  

Let $\mathcal{K} = \{1, \ldots, K\}$ be the set of classifiers. Let $\mathcal{N}_0 = \{1, \ldots, N_0\}$ and $\mathcal{N}_1 = \{1, \ldots, N_1\}$ be the sets of negative and positive data points respectively, with $N = N_0 + N_1$ being the total number of data points. Consider a $N_1 \times K$ matrix $B$ where $\beta_{ik} = 1$ if classifier $k \in \mathcal{K}$ correctly classified data point $i \in \mathcal{N}_1$ as positive, $\beta_{ik} = 0$ if it mistakenly classified $i$ as negative. Accordingly consider a $N_0 \times K$ matrix $A$ where $\alpha_{jk} = 0$ if classifier $k \in \mathcal{K}$ correctly classified data point $j \in \mathcal{N}_0$ as negative, $\alpha_{jk} = 1$ if $j$ was mistakenly classified as positive. 

Suppose $\mathcal{S} \subseteq \mathcal{K}$ is a set of $S$ classifiers selected to compose a given pruned ensemble. For any data point $i \in \mathcal{N}_1, \sum_{s \in \mathcal{S}} \beta_{is}$ is the number of correct positive classifications within $\mathcal{S}$. Accordingly, for any data point $j \in \mathcal{N}_0, \sum_{s \in \mathcal{S}} \alpha_{js}$ represents the number of (wrong) positive classifications within $\mathcal{S}$.

We define a threshold $0 \leq L \leq S$ such that for a given data point $i \in \mathcal{N}_1$, $\sum_{s \in \mathcal{S}} \beta_{is} > L$ implies that the ensemble classifies $i$ as positive. If $\sum_{s \in \mathcal{S}} \beta_{is} \leq L$, then $i$ is classified by the ensemble as negative. Similarly for $j \in \mathcal{N}_0$, $\sum_{s \in \mathcal{S}} \alpha_{js} > L$ implies a positive ensemble classification and $\sum_{s \in \mathcal{S}} \alpha_{js} \leq L$ implies a negative ensemble classification. For instance, if $S = 10$ and $L = 5$, then the ensemble classifies a data point as positive if at least 6 individual classifications are positive. If 5 or less are positive, then the ensemble classifies that data point as negative.

In our formulation we let the optimisation define both $\mathcal{S}$ and $L$. Hence we include $L$ as a general integer variable representing the classification threshold and binary variables $x_k = 1$ if classifier $k \in \mathcal{K}$ is chosen to compose the ensemble ($x_k = 0$ otherwise). 

\begin{table}[ht]
\centering
{
\renewcommand{\tabcolsep}{1.8mm} 
\renewcommand{\arraystretch}{1.4} 
 \begin{tabular}{cccc}
 & & \multicolumn{2}{c}{Predicted}\\
 & & \multicolumn{1}{c}{1} & \multicolumn{1}{c}{0} \\
 \cline{3-4}
 \parbox[t]{0mm}{\multirow{3}{*}{\rotatebox[origin=c]{90}{\hspace{0.4cm}Actual}}}$\quad$ & 
                 1 & \multicolumn{1}{|c}{$T^{+}$} & \multicolumn{1}{c|}{$F^{-}$}\\
 & 0 & \multicolumn{1}{|c}{$F^{+}$} & \multicolumn{1}{c|}{$T^{-}$} \\
 \cline{3-4}
 \end{tabular}
 \caption{Binary classification confusion matrix}
 \label{tab:cMatrix}
}
\end{table}

Consider the binary confusion matrix given in Table \ref{tab:cMatrix}, where $T^+, F^-, T^-$ and $F^+$ are the total number of classifications of each possible pattern. For each patterns we assign weights $W_T^+, W_T^-, W_F^+, W_F^- \in \mathbb{R}$, and the objective function is defined by the weighted sum $W_T^+ T^+ + W_F^- F^- +  W_T^- T^- + W_F^+ F^+$.



For modelling this weighted sum we define binary variables $t^+_i, f^-_i$ if the ensemble classification of $i \in \mathcal{N}_1$ is respectively a true positive or false negative. Similarly we define binary variables $t^-_j, f^+_j$ if the ensemble classification of $j \in \mathcal{N}_0$ is a true negative or false positive. The IP formulation that optimises a weighted sum of the patterns in the binary confusion matrix is given below:
{\small
\begin{equation}
\max  \sum_{i = 1}^{N_1} (W_T^+ t^{+}_i + W_F^-f^{-}_i) +  \sum_{j = 1}^{N_0} (W_T^- t^{-}_j + W_F^+ f^{+}_j)
\label{miobjective}
\end{equation}
}
\noindent subject to
\vspace{-0.2cm}
{\small
\begin{align}
(L+1) - \sum_{k = 1}^K x_k \;\beta_{ik} & \leq (K+1) (1 - t^+_i), & \forall i \in \mathcal{N}_1 \label{mi1} \\
\sum_{k = 1}^K x_k \;\beta_{ik} - L & \leq (K+1)t^+_i, & \forall i \in \mathcal{N}_1 \label{mi2} \\
t^+_i + f^-_i  & = 1, & \forall i  \in \mathcal{N}_1 \label{mi3}\\
\sum_{k = 1}^K x_k \; \alpha_{jk} - L & \leq K (1 - t^-_j), & \forall j \in \mathcal{N}_0 \label{mi4}\\
(L+1) - \sum_{k = 1}^K x_k \; \alpha_{jk}  & \leq K t^-_j, & \forall j \in \mathcal{N}_0 \label{mi5}\\
f^+_j + t^-_j & = 1, & \forall j  \in \mathcal{N}_0\label{mi6}\\
x_k & \in \mathbb{B} & \forall k \in \cal{K} \label{mi7}
\end{align}
\begin{align}
t^+_i, f^-_i & \in \mathbb{B} & \forall i \in \mathcal{N}_1 \label{mi8}\\
t^-_j, f^+_j & \in \mathbb{B} & \forall j \in \mathcal{N}_0 \label{mi9}\\
0 \leq L & \leq K, \label{mi10}\\
L & \in \mathbb{Z}\label{mi11}
\end{align}
}

\vspace{-0.6cm}

Constraints (\ref{mi1}) ensure that a positive data point $i \in \mathcal{N}_1$ has $t^+_i = 1$ if the number of individual positive classifications exceeds $L$. Conversely, constraints (\ref{mi2}) ensure that $t^+_i = 0$ if the number of individual positive classifications is no more than $L$. Constraints (\ref{mi3}) ensure that either $t^+_i = 1$ or $f^-_i = 1$. Constraints (\ref{mi4}) guarantee that a negative data point $j \in \mathcal{N}_0$ has $t^-_j = 0$ if the number of positive classifications exceeds $L$. Otherwise, constraints (\ref{mi5}) make sure that $t^-_j =1$. Constraints (\ref{mi6}) ensure that either $f^+_j = 1$ or $t^-_j = 1$. Constraints (\ref{mi7}-\ref{mi11}) define variables bounds. 

\subsection{Objective function}
\label{sec:obj}

For some classification problems, it may be desirable to optimise some patterns instead of others. For instance, in an investment decision, investing in the wrong project may cause bankruptcy while not investing in a promising project may be seen as a regretful but acceptable lost opportunity. In this case prioritising the minimisation of $F^+$ is desirable. The weights in Equation (\ref{miobjective}) provide flexibility for defining optimisation criteria depending on the characteristics of the dataset at hand (such as being highly imbalanced). A few examples are outlined below.

Accuracy is defined as $\frac{T^+ + T^-}{N}$. As $N$ is constant we can maximise accuracy by defining weights $W_T^+ = W_T^- = 1$ and $W_F^+ = W_F^- = 0$. Notice that if we choose this objective then constraints (\ref{mi2}) and (\ref{mi5}) are redundant as maximising positive weights $W_T^+$ and $W_T^-$ ensure that $t^+_i = 1$ and $t^-_i = 1$ if allowed by constraints (\ref{mi1}) and (\ref{mi4}). Similarly recall is defined as $\frac{T^+}{T^+ + F^-} = \frac{T^+}{N_1}$ and can be maximised by setting $W_T^+ = 1$ and $W_T^- = W_F^+ = W_F^- = 0$ (with constraints (\ref{mi2}) being redundant).

Now consider  $\theta = \frac{N_1}{N}$ as the dataset imbalance level. If, for instance, $\theta \leq 1 - \epsilon$ for small $\epsilon$, a high accuracy can be achieved by classifying every data point as positive. For imbalanced datasets a possibly useful configuration of objective function (\ref{miobjective}) is by setting weights $W_T^+ = (1 - \theta)$, $W_T^- = \theta$ and $W_F^+ = W_F^- = 0$. 

\section{Diversity}
\label{sec:diversity}

As mentioned before many ensemble pruning algorithms employ diversity criteria. Our proposed formulation optimises a performance measure, and in this section we introduce a way to control diversity with linear constraints. 

We consider a diversity measure called Pairwise Failure Crediting (PFC), proposed originally by \cite{chandra2006}, chosen due to satisfactory performance in imbalanced datasets \citep{bhowan2012,fernandes2015}. 
PFC measures how diverse an individual classifier is from the remaining classifiers in the ensemble. 

PFC is calculated as follows. For each classifier $k$, we compute a {\it failure pattern} (FP). A FP is a string of 0's and 1's with length $N$. A `0' in the string means that the classifier failed to correctly predict the corresponding data point and a `1' means that it predicted the data point correctly (irrespective of its real value). Once we have all failure patterns we take any two classifiers $k$ and $l$ and calculate their Hamming distance. The Hamming distance between same-length strings is the number of different characters in the same positions. For example, if $\text{FP}_k = \{0011011101\}$ and $\text{FP}_l = \{0110001110\}$, the Hamming distance between $k$ and $l$ is 5 (characters 2, 4, 6, 9 and 10 differ). Next, we sum all failures by both classifiers - that is, we sum the number of zeros in both strings which, in the example, is 9. The {\it failure credit} (FC) between $k$ and $l$ is obtained by dividing the Hamming distance by the sum of failures. In the example, $\text{FC}_{kl} = 5/9$. For every pair $k,l \in \mathcal{K}$ we compute $\text{FC}_{kl}$.

Consider again $\mathcal{S}$ as a set of $S \leq K$ classifiers that are selected to compose an ensemble. We assume without loss of generality that classifiers in $\mathcal{S}$ are indexed by $k = 1, \ldots, S$. Their PFC values are given by:
{\small
\begin{align*}
\text{PFC}_k & = \frac{\sum_{l = 1, l \neq k}^S \text{FC}_{kl} }{S - 1} & k \in \cal{S}
\end{align*}
}

\noindent A (maximum) value of 1 in $\text{PFC}_k$ means that $k$ classifies all data points differently from every other classifier in the ensemble, and a (minimum) value of 0 means that $k$ is identical to all other classifiers. Both extreme cases imply that all other classifiers are identical among themselves. 

For ensuring minimum desired diversity levels, we propose two approaches: (i) the minimum PFC value of any individual classifier is at least a certain threshold $0 \leq \tau \leq 1$ in order to prevent very similar pairs of classifiers and (ii) the average PFC value of the ensemble must be at least a certain threshold $0 \leq \gamma \leq 1$ to ensure an overall good level of diversity. Clearly we must have $\gamma \geq \tau$.

We add the following new decision variables. Let $y_{kl} = 1$ if both classifiers $k$ and $l$ have been selected to be part of the ensemble, and $y_{kl} = 0$ if at most one of $k$ and $l$ is chosen to compose the ensemble. This adds $\binom{K}{2}$ extra variables (for every possible pair $k,l$). For simplicity, both $y_{kl}$ and $y_{lk}$ denote the exact same variable. The following constraints ensure that $y_{kl}$ takes the correct values:
{\small
\begin{align}
y_{kl} & \geq x_k + x_l - 1 & \forall k, l \in \mathcal{K}, k < l  \label{mi12}\\
y_{kl}  & \leq x_k & \forall k, l \in \mathcal{K}, k < l \label{mi13}\\
y_{kl}  & \leq x_l & \forall  k, l \in \mathcal{K}, k < l \label{mi14}\\
y_{kl} & \geq 0 & \forall  k, l \in \mathcal{K}, k < l \label{mi15}
\end{align}
}

\noindent Notice that there is no need for the $y_{kl}$ variables to be binary. As both $x_k$ and $x_l$ are binary, $y_{kl}$ must have integer values in any integer solution. 

We can then rewrite the PFC equation using variables $x_k$ and $y_{kl}$:
{\small
\begin{align*}
\text{PFC}_k & = \frac{\sum_{l = 1, l \neq k}^K \text{FC}_{kl} \;y_{kl}}{\sum_{m = 1}^K x_m - 1}& \forall k \in \mathcal{K}
\end{align*}
}

\noindent The term $\sum_{m = 1}^K x_m$ is the cardinality of the ensemble and any non-selected classifier $k$ (with $x_k = 0$) has a PFC value of 0 (as all $y_{kl} = 0, l \neq k$).

The following linear constraints ensure that every classifier has a minimum PFC value of $\tau$:
{\small
\begin{align}
\sum_{\substack{l = 1 \\ l \neq k}}^K \text{FC}_{kl} \;y_{kl} & \geq \tau \Big(\sum_{m = 1}^K x_m - 1\Big) - K\tau (1 - x_k) & \forall k \in \mathcal{K} \label{eqpfc1}
\end{align}
}

\noindent The term $K \tau(1 - x_k)$ ensures that the constraints above are only enforced if classifier $k$ is chosen to compose the ensemble.

The following nonlinear constraint ensures that the average PFC of the ensemble is at least $\gamma$:
{\small
\begin{align}
\frac{1}{ \sum_{m = 1}^K x_m} \; \frac{\sum_{k = 1}^K \sum_{l = 1, l \neq k}^K \text{FC}_{kl} \;y_{kl}}{\sum_{m = 1}^K x_m - 1} & \geq \gamma
\label{eqnonlinear}
\end{align}
}

\noindent Observe that in Equation (\ref{eqnonlinear}) the FC value of every pair is added twice. We use this fact to linearise this expression. For a given subset $\mathcal{S}$, the average PFC $\mu_{\text{PFC}}$ is given by:
{\small
\begin{align*}
\mu_{\text{PFC}} & = \frac{1}{S} \sum_{k = 1}^S \frac{\sum_{l = 1, l \neq k}^S \text{FC}_{kl}}{S - 1} \\
                 & = \frac{1}{S(S-1)} \sum_{k = 1}^S \sum_{\substack{l = 1 \\ l \neq k}}^S \text{FC}_{kl} \\
                 & = \frac{2}{S(S-1)} \sum_{k = 1}^{S-1}\sum_{l = k + 1}^S \text{FC}_{kl} \\
                 & = \frac{1}{\binom{S}{2}} \sum_{k = 1}^{S-1}\sum_{l = k + 1}^S \text{FC}_{kl} = \mu_{\text{FC}}
\end{align*}
}

\noindent where $\mu_{\text{FC}}$ denotes the average FC value of all pairs in the ensemble. We conclude that $\mu_{FC} = \mu_{PFC}$, that is, the average PFC among all classifiers in the ensemble is equal to the average FC among all pairs. 

If $S$ classifiers are selected in the ensemble, then the number of $y_{kl}$ variables that take value 1 is exactly $\binom{S}{2}$. Therefore we can ensure that the average PFC value is at least $\gamma$ with the following linear constraint:
{\small
\begin{align}
\sum_{k = 1}^{K-1} \sum_{l = k+1}^{K} \text{FC}_{kl} \; y_{kl} & \geq \gamma \sum_{k = 1}^{K-1} \sum_{l = k+1}^{K}  y_{kl}
\label{eqpfc2}
\end{align}
}

The expanded formulation with minimum diversity levels is given by maximising (\ref{miobjective}) subject to (\ref{mi1})-(\ref{eqpfc1}) and (\ref{eqpfc2}). It requires $\binom{K}{2}$ extra variables and a similar number of extra constraints, which could lead it to be more computationally demanding. However we observed empirically in Section \ref{sec:solving} that the inclusion of such constraints causes a negligible decrease in solution quality.

\section{Computational experiments}
\label{sec:results}

In this section we outline the computational experiments used to evaluate the proposed formulation. We used 9 publicly available datasets, outlined in Table \ref{tab:ds}\footnote{All datasets used in this paper can be found at the UCI Machine Learning Repository \citep{lichman2013}}, and ranging from $N = 195$ to $N = 60000$. We also include a column with the value of the imbalance parameter $\theta$.

\subsection{Description of the experiments}
\label{sec:description}

We prepared 10 different heterogeneous classifier models. Each  model was instantiated a number of times with different random seeds and parameters. We set $K$ as multiples of 10 in order to have an equal number of instantiations of each classifier. For instance, if $K = 60$, we have 6 classifiers of each model. In our experiments, reported below, we used $K = \{40, 60, 80, 100\}$. Each classifier produces, as output, a probability of a data point being positive. This probability is rounded to define matrices $A$ and $B$. A more thorough description of the classifiers can be found in the supplementary material accompanying this paper.

For evaluating performance we used a stratified 10-fold cross-validation procedure. The $N$ data points are initially shuffled randomly and the dataset is split into 10 folds. At each iteration, one of the folds is left out as an independent set (unseen by the algorithms). The results presented below are based solely in this set. The other 9 folds, comprising 90\% of the original dataset, are joined and further split into two sets: a training set, containing 63\% of the data points, is used to optimise the individual classifiers. A validation set, comprising the remaining 27\% data points, is used to optimise the ensemble algorithms.

The procedure above is repeated 10 times: in each we vary the random seeds required to both shuffle the dataset and initialise the individual classifiers. For each value of $K$ and for each instance shown in Table \ref{tab:ds}, we run 100 experiments: 10 random initialisations $\times$ 10 folds. For ensuring reproducibility of our results, we have made all necessary data publicly available. A link can be found in the supplementary material.

\begin{table*}[ht]
\caption{Selected datasets from the UCI Machine Learning Repository \citep{lichman2013}}
\label{tab:ds}
\centering
\renewcommand{\tabcolsep}{1.8mm} 
\renewcommand{\arraystretch}{1.1} 
{\small
\begin{tabular}{l>{\raggedright\arraybackslash}p{4.2cm}ccccc}
\multicolumn{1}{c}{Identifier} & \multicolumn{1}{c}{Dataset name} & \multicolumn{1}{c}{Features} & \multicolumn{1}{c}{$N$} & \multicolumn{1}{c}{$N_0$} & \multicolumn{1}{c}{$N_1$} & \multicolumn{1}{c}{$\theta$}\\
\hline
PRK & Parkinsons & 23 & 195 & 48 & 147 & 0.77\\
MSK & Musk (Version 1) & 168 & 476 & 269 & 207 & 0.44\\ 
BCW & Breast Cancer Wisconsin & 32 & 569 & 357 & 212 & 0.37 \\
QSR & QSAR biodegradation & 41 & 1055 & 356 & 699 & 0.66 \\
DRD & Diabetic Retinopathy Debrecen & 20 & 1151 & 540 & 611 & 0.53 \\
SPA & Spambase & 57 & 4601 & 2788 & 1813 & 0.39 \\
DEF & Default of credit card clients & 24 & 30000 & 23364 & 6636 & 0.22 \\
BMK & Bank Marketing & 21 & 41188 & 36548 & 4640 & 0.11 \\
APS & APS Failure at Scania Trucks & 171 &  60000 & 59000 &  1000 & 0.02  
\end{tabular}
}
\end{table*}

\subsection{Benchmarks}

To evaluate our formulation, we compare it to seven other approaches: Full (non-pruned) Ensemble (FE), Reduced-Error Prunning with Backfitting \citep{friedman1981} (hereby Backfitting or BFT), Kappa pruning \citep{margineantu1997} and four different hill climbing based methods. Due to space constraints we report here only results for only four approaches. The full results are available in the supplementary material accompanying this paper. All benchmarks classify data points based on majority voting.

Backfitting aims to maximise accuracy by following a greedy approach with revision. Initially, the classifier subset $S$ is empty. At each iteration, the algorithm adds to $S$ a classifier $s$ such that the accuracy of $S \cup {s}$ is maximised. This process is repeated until a predefined number $M$ of classifiers is added to $S$. Ties are broken arbitrarily. Whenever a classifier is added, the greedy choice is revised through a local search procedure. Each classifier in the ensemble is iteratively replaced by another previously left out. If the overall accuracy is improved, the method starts again with the new subset $S$. The local search stops after 100 iterations or whenever no classifier is able to improve the solution, whichever happens first. Backfitting requires the ensemble subset size to be fixed. For a fairer comparison, we varied $M$ within 20\% and 80\% of $K$. The best in-sample results are used to evaluate the independent set. 

Kappa pruning follows a similar procedure, but with two notable differences: there is no revision of the greedy choice and it optimises the $\kappa$-statistic \citep{cohen1960}, a diversity measure of statistical agreement between any two classifiers. Both Backfitting and Kappa pruning require the ensemble subset size to be fixed. For a fairer comparison, we varied $M$ within 20\% and 80\% of $K$. The best in-sample results are used to evaluate the independent set. 

The other benchmarks use the forward version of the hill climbing search algorithm, and they differ only in the selected fitness function. In all four methods, the first iteration selects the individual classifier with maximum accuracy, regardless of the fitness function chosen. In the following iterations, classifiers are greedily added so as to maximise the selected fitness. This process is repeated until all classifiers are added to $S$. In the end, the chosen ensemble is the one with the best accuracy over all the ensembles iteratively created. As opposed to the other benchmarks, direct hill climbing does not define the ensemble size {\it a priori}. The fitness functions chosen are the same as tested by \cite{partalas2010}: Accuracy, Complementariness \citep{martinez2004}, Concurrency \citep{banfield2005} and Uncertainty Weighted Accuracy \citep{partalas2010}. We refer to them as HC-ACC, HC-CON, HC-COM and HC-UWA. 

In the experiments reported below we compare our formulation to BFT, two hill climbing methods, HC-CON and HC-UWA (which had a better performance when compared to HC-ACC and HC-COM) and FE (Full Ensemble). The complete set of results is reported in the supplementary material. All algorithms are allowed to run for a maximum of 5 minutes.

\subsection{Solving the formulation}
\label{sec:solving}

0-1 IP is notoriously computationally difficult. Its feasibility version is one of Karp's 21 NP-complete problems \citep{karp1972}. Modern day solvers however can effectively solve large instances of several different IP problems by employing a combination of techniques (such as branch-and-bound and cutting planes). From an optimisation point of view, important considerations are (i) up to what size are modern solvers able to optimally solve our formulation within a reasonable time limit, and (ii) if the limit is reached prior to proving optimality, how large are the optimality gaps and how quickly does the solver find ``good enough'' solutions? 

While searching for these answers and studying traditional IP techniques in order to improve computational performance is within the scope of our current and future work, we have opted, due to limited space, to refrain from further discussing them. We however observed in practice that, with a time limit of only 5 minutes, we were able to either optimally solve or terminate the algorithm with small optimality gaps for all instances. These gaps are shown in Table \ref{tableGaps}.

\begin{table}[ht]
\begin{center}
\caption{Average optimality gaps and corresponding standard deviations (in \%).}
\label{tableGaps}
\renewcommand{\tabcolsep}{1.8mm} 
\renewcommand{\arraystretch}{1.1} 
{\small
\begin{tabular}{|l|r|rr|rr|}
\hline
\multirow{2}{*}{Instance} & \multirow{2}{*}{27\% of $N$} & \multicolumn{2}{c|}{No diversity} & \multicolumn{2}{c|}{With diversity} \\
                          &                      & \multicolumn{1}{l}{Avg.}   & \multicolumn{1}{l|}{Std.}   & \multicolumn{1}{l}{Avg.}   & \multicolumn{1}{c|}{Std.}   \\ \hline
PRK                       & 26                   & 0.0                        & 0.0                         & 0.0                         & 0.0                         \\
BCW                       & 129                  & 0.0                        & 0.0                         & 0.0                         & 0.0                         \\
MSK                       & 154                  & 0.0                        & 0.0                         & 0.0                         & 0.0                         \\
QSR                       & 285                  & 0.0                        & 0.0                         & 0.0                         & 0.1                         \\
DRD                       & 311                  & 4.3                        & 2.0                         & 6.1                         & 1.8                         \\
SPA                       & 1242                 & 0.0                        & 0.0                         & 0.2                         & 0.2                         \\
DEF                       & 8100                 & 6.7                        & 0.4                         & 6.9                         & 0.4                        \\
BMK                       & 11121                & 5.4                        & 0.3                         & 5.5                         & 0.2                         \\
APS                       & 16200                & 0.1                        & 0.0                         & 0.1                         & 0.0                         \\
\hline
\end{tabular}
}
\end{center}
\end{table}

This table summarises the average gaps and their respective standard deviations for experiments reported below in Table \ref{table1} with $K = 100$. The ``No diversity'' column corresponds to {\bf F1} in that table, and only constraints (\ref{mi1})-(\ref{mi11}) are used.  The ``With diversity'' column corresponds to {\bf F3}, which uses constraints (\ref{mi1})-(\ref{eqpfc1}) and (\ref{eqpfc2}). The largest instance, APS, had average gaps of only 0.1\% in both cases. The hardest instance was DEF (6.7\% and 6.9\%). The only case where a difference was notable was for the DRD instance (4.3\% and 6.1\%).


In our view, even the hardest instances were still relatively close to optimality considering the short computational time. For this reason, in all computational experiments reported below, we set a time limit of 5 minutes. If the time limit is reached, we halt the solver and retrieve the best solution available at that point. We employed CPLEX 12.8 \citep{cplex128} with default parameters as the IP solver and we run all experiments in an Intel Core(TM) I7-7700 @ 3.60GHz with 32GB of RAM, using 8 cores and having Linux as the operating system.

\subsection{Accuracy}
\label{sec:resultsAccuracy}

In our first suite of experiments we set weights $W_T^+ = W_T^- = 1$ and $W_F^+ = W_F^- = 0$, that is, we seek to maximise classification accuracy regardless of the dataset imbalance level. To evaluate both the formulation introduced in Section \ref{sec:model} and the diversity constraints introduced in Section \ref{sec:diversity}, we propose three different configurations.

In the first we employ only constraints (\ref{mi1})-(\ref{mi11}), without enforcing diversity - we refer to this configuration as {\bf F1}. The other two configurations, {\bf F2} and {\bf F3}, enforce minimum diversity levels in the hope of preventing possible overfitting of the training and validation sets. In {\bf F2} we set $\tau = 0$ and $\gamma = \frac{\text{PFC}_{\text{min}} + \text{PFC}_{\text{avg}}}{2}$, where $\text{PFC}_{\text{min}}$ and $\text{PFC}_{\text{avg}}$ are the minimum individual PFC among all classifiers and the average PFC value of the full ensemble. So in {\bf F2} we only constrain the overall average PFC value. In {\bf F3} we also set $\tau = \text{PFC}_{\text{min}}$, so restricting individual PFC values as well as the average PFC.

\begin{table}[ht]
\centering
\caption{Out-of-sample average accuraries and standard deviations.}
\label{table1}
\renewcommand{\tabcolsep}{1mm} 
\renewcommand{\arraystretch}{1.4} 
{\scriptsize
\begin{tabular}{|lr|rr|rr|rr|rr|rr|rr|rr|}
\hline
\multirow{2}{*}{Dataset} & \multirow{2}{*}{$K$} & \multicolumn{2}{c|}{\textbf{F1}} & \multicolumn{2}{c|}{\textbf{F2}} & \multicolumn{2}{c|}{\textbf{F3}} & \multicolumn{2}{c|}{BFT} & \multicolumn{2}{c|}{HC-CON} & \multicolumn{2}{c|}{HC-UWA} & \multicolumn{2}{c|}{FE} \\
 & & \multicolumn{1}{c}{Avg.} & \multicolumn{1}{c|}{Std.} & \multicolumn{1}{c}{Avg.} & \multicolumn{1}{c|}{Std.} & \multicolumn{1}{c}{Avg.} & \multicolumn{1}{c|}{Std.} & \multicolumn{1}{c}{Avg.} & \multicolumn{1}{c|}{Std.} & \multicolumn{1}{c}{Avg.} & \multicolumn{1}{c|}{Std.} & \multicolumn{1}{c}{Avg.} & \multicolumn{1}{c|}{Std.} & \multicolumn{1}{c}{Avg.} & Std. \\ \hline
PRK &  40 & 0.9066 & 0.0145 & \textbf{0.9134} & 0.0136 & 0.9114 & 0.0130 & 0.9041 & 0.0156 & 0.9118 & 0.0152 & 0.9098 & 0.0183 & 0.9014 & 0.0066 \\
    &  60 & 0.9036 & 0.0124 & \textbf{0.9089} & 0.0103 & 0.9084 & 0.0113 & 0.8999 & 0.0146 & 0.9083 & 0.0176 & 0.9089 & 0.0195 & 0.8994 & 0.0088 \\
    &  80 & 0.9051 & 0.0144 & 0.9052 & 0.0124 & 0.9056 & 0.0131 & 0.9015 & 0.0084 & 0.9048 & 0.0167 & 0.9129 & 0.0223 & 0.9004 & 0.0091 \\
    & 100 & 0.9036 & 0.0154 & 0.9070 & 0.0202 & 0.9042 & 0.0183 & 0.9020 & 0.0080 & 0.9089 & 0.0174 & 0.9120 & 0.0213 & 0.8959 & 0.0092 \\ \hline
BCW &  40 & 0.9664 & 0.0051 & \textbf{0.9689} & 0.0052 & \textbf{0.9690} & 0.0054 & 0.9652 & 0.0058 & 0.9680 & 0.0032 & 0.9671 & 0.0050 & 0.9632 & 0.0058 \\
    &  60 & 0.9676 & 0.0043 & \textbf{0.9701} & 0.0036 & \textbf{0.9703} & 0.0038 & 0.9671 & 0.0056 & 0.9696 & 0.0045 & 0.9680 & 0.0040 & 0.9629 & 0.0056 \\
    &  80 & 0.9643 & 0.0057 & 0.9673 & 0.0032 & 0.9673 & 0.0032 & 0.9660 & 0.0049 & 0.9710 & 0.0048 & 0.9701 & 0.0049 & 0.9629 & 0.0051 \\
    & 100 & 0.9618 & 0.0052 & 0.9676 & 0.0036 & 0.9676 & 0.0040 & 0.9673 & 0.0047 & 0.9710 & 0.0052 & 0.9692 & 0.0049 & 0.9627 & 0.0056 \\ \hline
MSK & 40 & 0.9109 & 0.0110 & 0.9165 & 0.0093 & 0.9127 & 0.0090 & 0.9117 & 0.0122 & 0.9238 & 0.0092 & 0.9192 & 0.0126 & 0.8896 & 0.0100 \\
    & 60 & 0.9119 & 0.0086 & 0.9129 & 0.0092 & 0.9115 & 0.0091 & 0.9083 & 0.0136 & 0.9230 & 0.0103 & 0.9177 & 0.0114 & 0.8869 & 0.0097 \\
    & 80 & 0.9192 & 0.0097 & 0.9208 & 0.0078 & 0.9208 & 0.0076 & 0.9163 & 0.0117 & 0.9291 & 0.0090 & 0.9207 & 0.0096 & 0.8894 & 0.0099 \\
    & 100 & 0.9196 & 0.0066 & 0.9192 & 0.0075 & 0.9188 & 0.0085 & 0.9098 & 0.0135 & 0.9299 & 0.0081 & 0.9215 & 0.0077 & 0.8885 & 0.0096 \\ \hline
QSR & 40 & \textbf{0.8740} & 0.0041 & 0.8731 & 0.0038 & \textbf{0.8742} & 0.0047 & 0.8719 & 0.0046 & 0.8734 & 0.0049 & 0.8733 & 0.0040 & 0.8706 & 0.0033 \\
    & 60 & \textbf{0.8755} & 0.0056 & \textbf{0.8756} & 0.0045 & \textbf{0.8740} & 0.0055 & 0.8710 & 0.0050 & 0.8736 & 0.0077 & 0.8725 & 0.0051 & 0.8692 & 0.0039 \\
    & 80 & 0.8715 & 0.0047 & 0.8728 & 0.0047 & 0.8740 & 0.0033 & 0.8747 & 0.0045 & 0.8731 & 0.0062 & 0.8747 & 0.0055 & 0.8706 & 0.0025 \\
    & 100 & 0.8722 & 0.0054 & \textbf{0.8757} & 0.0027 & 0.8732 & 0.0042 & 0.8729 & 0.0050 & 0.8724 & 0.0040 & 0.8738 & 0.0060 & 0.8705 & 0.0028 \\ \hline
DRD & 40 & 0.7401 & 0.0060 & 0.7405 & 0.0091 & 0.7389 & 0.0075 & 0.7415 & 0.0080 & 0.7418 & 0.0077 & 0.7467 & 0.0070 & 0.7086 & 0.0080 \\
    & 60 & 0.7458 & 0.0114 & 0.7462 & 0.0096 & 0.7448 & 0.0104 & 0.7475 & 0.0100 & 0.7480 & 0.0086 & 0.7497 & 0.0068 & 0.7110 & 0.0077 \\
    & 80 & 0.7449 & 0.0063 & 0.7412 & 0.0054 & 0.7437 & 0.0068 & 0.7464 & 0.0060 & 0.7501 & 0.0065 & 0.7492 & 0.0070 & 0.7094 & 0.0078 \\
    & 100 & 0.7401 & 0.0083 & 0.7459 & 0.0054 & 0.7442 & 0.0065 & 0.7441 & 0.0065 & 0.7486 & 0.0079 & 0.7516 & 0.0056 & 0.7114 & 0.0076 \\ \hline
SPA & 40 & 0.9529 & 0.0014 & 0.9534 & 0.0013 & 0.9534 & 0.0013 & 0.9529 & 0.0008 & 0.9535 & 0.0012 & 0.9523 & 0.0011 & 0.9476 & 0.0010 \\
    & 60 & 0.9533 & 0.0013 & 0.9542 & 0.0015 & 0.9537 & 0.0017 & 0.9535 & 0.0014 & 0.9547 & 0.0014 & 0.9534 & 0.0012 & 0.9469 & 0.0007 \\
    & 80 & 0.9538 & 0.0018 & 0.9536 & 0.0020 & 0.9536 & 0.0014 & 0.9532 & 0.0014 & 0.9554 & 0.0016 & 0.9536 & 0.0018 & 0.9468 & 0.0008 \\
    & 100 & 0.9535 & 0.0018 & 0.9538 & 0.0016 & 0.9537 & 0.0016 & 0.9530 & 0.0014 & 0.9557 & 0.0012 & 0.9537 & 0.0015 & 0.9467 & 0.0006 \\ \hline
DEF & 40 & 0.8206 & 0.0005 & 0.8205 & 0.0005 & 0.8204 & 0.0005 & 0.8206 & 0.0004 & 0.8205 & 0.0005 & 0.8211 & 0.0004 & 0.8207 & 0.0004 \\
    & 60 & 0.8205 & 0.0005 & 0.8204 & 0.0005 & 0.8206 & 0.0003 & 0.8207 & 0.0003 & 0.8204 & 0.0004 & 0.8211 & 0.0004 & 0.8205 & 0.0006 \\
    & 80 & 0.8203 & 0.0006 & 0.8206 & 0.0006 & 0.8206 & 0.0008 & 0.8203 & 0.0004 & 0.8206 & 0.0004 & 0.8212 & 0.0003 & 0.8203 & 0.0004 \\
    & 100 & 0.8202 & 0.0004 & 0.8200 & 0.0003 & 0.8205 & 0.0006 & 0.8206 & 0.0004 & 0.8205 & 0.0004 & 0.8213 & 0.0003 & 0.8201 & 0.0004 \\ \hline
BMK & 40 & \textbf{0.9178} & 0.0004 & 0.9171 & 0.0006 & 0.9170 & 0.0006 & 0.9166 & 0.0006 & 0.9168 & 0.0007 & 0.9173 & 0.0005 & 0.9143 & 0.0007 \\
    & 60 & \textbf{0.9175} & 0.0005 & \textbf{0.9173} & 0.0006 & \textbf{0.9173} & 0.0006 & 0.9167 & 0.0003 & 0.9170 & 0.0007 & 0.9173 & 0.0005 & 0.9134 & 0.0007 \\
    & 80 & \textbf{0.9174} & 0.0007 & \textbf{0.9174} & 0.0007 & 0.9172 & 0.0005 & 0.9165 & 0.0006 & 0.9169 & 0.0009 & 0.9173 & 0.0006 & 0.9131 & 0.0005 \\
    & 100 & \textbf{0.9176} & 0.0004 & \textbf{0.9174} & 0.0005 & 0.9173 & 0.0007 & 0.9167 & 0.0004 & 0.9168 & 0.0007 & 0.9173 & 0.0008 & 0.9124 & 0.0006 \\ \hline
APS & 40 & 0.9937 & 0.0002 & 0.9938 & 0.0002 & 0.9938 & 0.0002 & 0.9936 & 0.0002 & 0.9938 & 0.0001 & 0.9938 & 0.0002 & 0.9929 & 0.0001 \\
    & 60 & \textbf{0.9938} & 0.0002 & \textbf{0.9938} & 0.0002 & \textbf{0.9938} & 0.0002 & 0.9936 & 0.0002 & 0.9937 & 0.0001 & 0.9938 & 0.0001 & 0.9925 & 0.0001 \\
    & 80 & \textbf{0.9939} & 0.0001 & 0.9938 & 0.0001 & 0.9938 & 0.0001 & 0.9936 & 0.0002 & 0.9938 & 0.0001 & 0.9937 & 0.0002 & 0.9925 & 0.0001 \\
    & 100 & 0.9938 & 0.0001 & \textbf{0.9939} & 0.0001 & \textbf{0.9939} & 0.0001 & 0.9937 & 0.0002 & 0.9938 & 0.0001 & 0.9938 & 0.0001 & 0.9925 & 0.0001 \\ \hline
\multicolumn{2}{r}{\textbf{Average:}} & \multicolumn{1}{r}{\textbf{0.8985}} & \multicolumn{1}{r}{\textbf{0.0049}} & \multicolumn{1}{r}{\textbf{0.8997}} & \multicolumn{1}{r}{\textbf{0.0045}} & \multicolumn{1}{r}{\textbf{0.8993}} & \multicolumn{1}{r}{\textbf{0.0046}} & \multicolumn{1}{r}{\textbf{0.8979}} & \multicolumn{1}{r}{\textbf{0.0049}} & \multicolumn{1}{r}{\textbf{0.9012}} & \multicolumn{1}{r}{\textbf{0.0052}} & \multicolumn{1}{r}{\textbf{0.9009}} & \multicolumn{1}{r}{\textbf{0.0055}} & \multicolumn{1}{r}{\textbf{0.8894}} & \multicolumn{1}{r}{\textbf{0.0041}}
\end{tabular}
}
\end{table}

Table \ref{table1} presents the results for the 9 datasets and four different values of $K$. For each algorithm, we include two columns: the average ({\bf Avg.}) and standard deviation ({\bf Std.}) of the out-of-sample accuracy, calculated with the data points in the independent set. We remind the reader that each entry in the table represents the average accuracy of 100 different runs. A bold value in any of the first three {\bf Avg.} columns means that our formulation obtained a higher average than all four benchmarks. The last row gives an overall average value of the corresponding column. We also present, in Table \ref{tableRanking1}, the average rank per value of $K$ across all datasets. The ranking procedure \citep{demvsar2006} works as follows. For each dataset, the best performing algorithm gets rank 1.0, the second best gets rank 2.0, and so on. If multiple algorithms tie, they are assigned the average of their ranks.

\begin{table}[ht]
\centering
\caption{Average ranks of accuracies}
\label{tableRanking1}
\renewcommand{\tabcolsep}{1.8mm} 
\renewcommand{\arraystretch}{1.4} 
{\scriptsize
\begin{tabular}{|l|rrr|rrrr|}
\hline
\multicolumn{1}{|c|}{$K$} & \multicolumn{1}{c}{\textbf{F1}} & \multicolumn{1}{c}{\textbf{F2}} & \multicolumn{1}{c|}{\textbf{F3}} & \multicolumn{1}{c}{BFT} & \multicolumn{1}{c}{HC-CON} & \multicolumn{1}{c}{HC-UWA} & \multicolumn{1}{c|}{FE} \\
\hline
40       &           3.82 &           3.74 &           3.79 &           4.04 &           3.71 &           3.66 &           5.25 \\
60       &           3.82 &           3.70 &           3.76 &           4.03 &           3.66 &           3.66 &           5.39 \\
80       &           3.82 &           3.80 &           3.75 &           4.09 &           3.54 &           3.58 &           5.42 \\
100      &           3.91 &           3.72 &           3.72 &           4.07 &           3.55 &           3.56 &           5.48 \\\hline
\multicolumn{1}{r}{\textbf{Avg:}} &  \multicolumn{1}{r}{\textbf{3.84}} &  \multicolumn{1}{r}{\textbf{3.74}} &  \multicolumn{1}{r}{\textbf{3.75}} &  \multicolumn{1}{r}{\textbf{4.06}} &  \multicolumn{1}{r}{\textbf{3.61}} &  \multicolumn{1}{r}{\textbf{3.61}} &  \multicolumn{1}{r}{\textbf{5.38}} \\
\end{tabular}
}
\end{table}

All methods in the tables have, as primary objective, maximising accuracy. The results suggest that while the formulation is overall competitive, it was slightly outperformed by HC-CON and HC-UWA - both in terms of average accuracy and average rank. Still, with the exception of FE, the difference between BFT (worst performing) and HC-CON (best performing) was 0.33\% in terms of average overall accuracy and 0.45 in terms of average rank. Adding diversity constraints to our formulation also had a small beneficial impact in improving average accuracy and reducing the average ranking. In 11 out of the 36 cases, \textbf{F2} outperformed all benchmarks. 

Both HC methods had  a higher dispersion of accuracies than BFT and our formulation. Moreover adding diversity in \textbf{F2} and \textbf{F3} helped reduce that dispersion. While adding diversity resulted in a small improvement, further studies on either better enforcing these constraints or proposing new constraints based on alternative diversity measures remain as future work. Since our proposed method is exact in nature (although limited to 5 minutes), in the supplementary material we discuss in more details the effects of overfitting.

\subsection{Balanced accuracy}

Several of the datasets used in this work are imbalanced and so accuracy may not be the best comparison metric among the methods. In this section, we use an alternative metric called Balanced Accuracy (BA), which weighs equally the accuracy of positive data points and the accuracy of negative data points. BA is a more appropriate measure for imbalanced datasets \citep{brodersen2010} and is given by:
{\small
\begin{align}
    \text{BA} = \frac{\frac{T^+}{T^+ + F^-} + \frac{T^-}{T^- + F^+}}{2} = \frac{\frac{T^+}{N_1} + \frac{T^-}{N_0}}{2}
    \label{eq:ba}
\end{align}
}

With regards to our formulation we analyse the performance of two different weights assignments in (\ref{miobjective}). Here we do not use any diversity constraints - we employ {\bf F1} as defined earlier and a modified {\bf F1} where we maximise the $\theta$-weighted configuration suggested in Section \ref{sec:obj}. In fact, Theorem \ref{theorem1} shows that maximising BA is equivalent to maximising the $\theta$-weighted function.

\begin{theorem}
Maximising the $\theta$-weighted configuration is equivalent to maximising balanced accuracy.
\label{theorem1}
\end{theorem}

\begin{proof}
Following the definition of the $\theta$-weighted function in Section \ref{sec:obj}, objective function $z$ can be written as:
{\small
\begin{align*}
\max z  & = \bigg(1 - \frac{N_1}{N}\bigg) T^+ + \frac{N_1}{N} T^-
\end{align*}
}

\noindent where {\small $T^+ = \sum_{i = 1}^{N_1} t^+_i$, $T^- = \sum_{j = 1}^{N_0} t^-_j$} and {\small $\theta = \frac{N_1}{N}$}. As {\small $N = N_0 + N_1$} it follows that:
{\small
\begin{align*}
\max \;\;\;z  & = \frac{  N_0}{N} T^+ + \frac{N_1}{N} T^-\\
  \max Nz & = N_0 T^+ + N_1 T^- \\
  \max \;\,cz & = \frac{T^+}{N_1} + \frac{T^-}{N_0}
\end{align*}
}

\noindent where constant $c = \frac{N}{N_1N_0} > 0$. If $c > 0$, the optimal solution does not depend on it since it is only a scaling factor in the solution value. That is valid also for $c = 2$ as in (\ref{eq:ba}), and thus maximising the $\theta$-weighted function is equivalent to maximising balanced accuracy.
\end{proof}

Tables \ref{table2} and \ref{tableRanking2}, which have the same structure as Tables \ref{table1} and \ref{tableRanking1}, show the results. We did not rerun the experiments for the accuracy version of {\bf F1} nor for the benchmarks, rather we used the same ensemble subsets to calculate the corresponding balanced accuracies.

\begin{table}[ht]
\centering
\caption{Balanced Accuracy averages and standard deviations}
\label{table2}
\renewcommand{\tabcolsep}{1.8mm} 
\renewcommand{\arraystretch}{1.4} 
{\scriptsize
\begin{tabular}{|lr|rr|rr|rr|rr|rr|rr|}
\hline
\multirow{2}{*}{Dataset} & \multirow{2}{*}{$K$} & \multicolumn{2}{c|}{\textbf{F1}} & \multicolumn{2}{c|}{\textbf{F1} ($\theta$-weighted)} & \multicolumn{2}{c|}{BFT} & \multicolumn{2}{c|}{HC-CON} & \multicolumn{2}{c|}{HC-UWA} & \multicolumn{2}{c|}{FE} \\
 & & \multicolumn{1}{c}{Avg.} & \multicolumn{1}{c|}{Std.} & \multicolumn{1}{c}{Avg.} & \multicolumn{1}{c|}{Std.} & \multicolumn{1}{c}{Avg.} & \multicolumn{1}{c|}{Std.} & \multicolumn{1}{c}{Avg.} & \multicolumn{1}{c|}{Std.} & \multicolumn{1}{c}{Avg.} & \multicolumn{1}{c|}{Std.} & \multicolumn{1}{c}{Avg.} & Std. \\ \hline
PRK & 40 & 0.8478 & 0.0328 & 0.8582 & 0.0289 & 0.8523 & 0.0177 & 0.8643 & 0.0286 & 0.8548 & 0.0290 & 0.8308 & 0.0110 \\
 & 60 & 0.8467 & 0.0242 & 0.8545 & 0.0181 & 0.8364 & 0.0242 & 0.8611 & 0.0309 & 0.8552 & 0.0310 & 0.8237 & 0.0158 \\
 & 80 & 0.8464 & 0.0290 & 0.8514 & 0.0204 & 0.8433 & 0.0169 & 0.8547 & 0.0244 & 0.8615 & 0.0331 & 0.8264 & 0.0146 \\
 & 100 & 0.8487 & 0.0306 & 0.8557 & 0.0266 & 0.8416 & 0.0103 & 0.8599 & 0.0283 & 0.8598 & 0.0320 & 0.8154 & 0.0120 \\\hline
BCW & 40 & \textbf{0.9645} & 0.0057 & \textbf{0.9668} & 0.0050 & 0.9581 & 0.0066 & 0.9624 & 0.0040 & 0.9608 & 0.0056 & 0.9566 & 0.0077 \\
 & 60 & \textbf{0.9666} & 0.0052 & \textbf{0.9675} & 0.0037 & 0.9608 & 0.0064 & 0.9638 & 0.0047 & 0.9620 & 0.0041 & 0.9564 & 0.0073 \\
 & 80 & 0.9646 & 0.0054 & 0.9651 & 0.0044 & 0.9601 & 0.0059 & 0.9655 & 0.0053 & 0.9647 & 0.0059 & 0.9562 & 0.0069 \\
 & 100 & 0.9625 & 0.0050 & 0.9631 & 0.0056 & 0.9616 & 0.0059 & 0.9658 & 0.0052 & 0.9642 & 0.0055 & 0.9562 & 0.0073 \\\hline
MSK & 40 & 0.9103 & 0.0102 & 0.9104 & 0.0083 & 0.9071 & 0.0133 & 0.9216 & 0.0090 & 0.9167 & 0.0129 & 0.8844 & 0.0108 \\
 & 60 & 0.9108 & 0.0080 & 0.9094 & 0.0084 & 0.9047 & 0.0143 & 0.9207 & 0.0103 & 0.9153 & 0.0124 & 0.8815 & 0.0102 \\
 & 80 & 0.9179 & 0.0100 & 0.9164 & 0.0075 & 0.9121 & 0.0124 & 0.9259 & 0.0095 & 0.9176 & 0.0101 & 0.8841 & 0.0105 \\
 & 100 & 0.9180 & 0.0071 & 0.9162 & 0.0058 & 0.9056 & 0.0143 & 0.9268 & 0.0083 & 0.9184 & 0.0082 & 0.8833 & 0.0102 \\\hline
QSR & 40 & \textbf{0.8574} & 0.0066 & \textbf{0.8645} & 0.0049 & 0.8496 & 0.0052 & 0.8533 & 0.0063 & 0.8505 & 0.0044 & 0.8432 & 0.0037 \\
 & 60 & \textbf{0.8604} & 0.0072 & \textbf{0.8639} & 0.0052 & 0.8467 & 0.0051 & 0.8533 & 0.0101 & 0.8503 & 0.0057 & 0.8407 & 0.0040 \\
 & 80 & \textbf{0.8559} & 0.0046 & \textbf{0.8649} & 0.0038 & 0.8515 & 0.0050 & 0.8532 & 0.0072 & 0.8530 & 0.0059 & 0.8421 & 0.0025 \\
 & 100 & \textbf{0.8563} & 0.0050 & \textbf{0.8648} & 0.0052 & 0.8477 & 0.0049 & 0.8530 & 0.0048 & 0.8521 & 0.0063 & 0.8419 & 0.0030 \\\hline
DRD & 40 & 0.7407 & 0.0056 & 0.7434 & 0.0062 & 0.7465 & 0.0078 & 0.7443 & 0.0080 & 0.7499 & 0.0073 & 0.7117 & 0.0079 \\
 & 60 & 0.7463 & 0.0115 & 0.7437 & 0.0088 & 0.7520 & 0.0098 & 0.7508 & 0.0089 & 0.7531 & 0.0068 & 0.7145 & 0.0075 \\
 & 80 & 0.7457 & 0.0064 & 0.7468 & 0.0046 & 0.7503 & 0.0057 & 0.7525 & 0.0065 & 0.7525 & 0.0072 & 0.7126 & 0.0077 \\
 & 100 & 0.7404 & 0.0088 & 0.7466 & 0.0093 & 0.7479 & 0.0063 & 0.7506 & 0.0075 & 0.7551 & 0.0057 & 0.7148 & 0.0075 \\\hline
SPA & 40 & \textbf{0.9493} & 0.0018 & \textbf{0.9513} & 0.0023 & 0.9478 & 0.0012 & 0.9490 & 0.0015 & 0.9478 & 0.0015 & 0.9415 & 0.0013 \\
 & 60 & 0.9498 & 0.0016 & \textbf{0.9514} & 0.0020 & 0.9489 & 0.0018 & 0.9503 & 0.0015 & 0.9491 & 0.0015 & 0.9405 & 0.0008 \\
 & 80 & 0.9504 & 0.0021 & \textbf{0.9521} & 0.0018 & 0.9488 & 0.0017 & 0.9512 & 0.0017 & 0.9493 & 0.0020 & 0.9403 & 0.0009 \\
 & 100 & 0.9500 & 0.0022 & \textbf{0.9517} & 0.0025 & 0.9485 & 0.0020 & 0.9515 & 0.0015 & 0.9493 & 0.0018 & 0.9402 & 0.0008 \\\hline
DEF & 40 & \textbf{0.6633} & 0.0022 & \textbf{0.6949} & 0.0023 & 0.6507 & 0.0011 & 0.6543 & 0.0024 & 0.6559 & 0.0009 & 0.6507 & 0.0011 \\
 & 60 & \textbf{0.6643} & 0.0020 & \textbf{0.6973} & 0.0021 & 0.6491 & 0.0018 & 0.6541 & 0.0017 & 0.6556 & 0.0009 & 0.6490 & 0.0012 \\
 & 80 & \textbf{0.6662} & 0.0014 & \textbf{0.6985} & 0.0016 & 0.6484 & 0.0020 & 0.6553 & 0.0009 & 0.6557 & 0.0011 & 0.6484 & 0.0011 \\
 & 100 & \textbf{0.6661} & 0.0019 & \textbf{0.6992} & 0.0014 & 0.6482 & 0.0018 & 0.6542 & 0.0020 & 0.6560 & 0.0007 & 0.6473 & 0.0012 \\\hline
BMK & 40 & \textbf{0.7731} & 0.0055 & \textbf{0.8607} & 0.0015 & 0.7331 & 0.0038 & 0.7453 & 0.0027 & 0.7462 & 0.0025 & 0.6993 & 0.0039 \\
 & 60 & \textbf{0.7763} & 0.0049 & \textbf{0.8662} & 0.0019 & 0.7338 & 0.0028 & 0.7477 & 0.0029 & 0.7469 & 0.0019 & 0.6883 & 0.0038 \\
 & 80 & \textbf{0.7765} & 0.0055 & \textbf{0.8684} & 0.0015 & 0.7322 & 0.0046 & 0.7477 & 0.0028 & 0.7469 & 0.0018 & 0.6822 & 0.0036 \\
 & 100 & \textbf{0.7794} & 0.0053 & \textbf{0.8694} & 0.0012 & 0.7363 & 0.0040 & 0.7478 & 0.0029 & 0.7479 & 0.0018 & 0.6762 & 0.0031 \\\hline
APS & 40 & \textbf{0.8690} & 0.0062 & \textbf{0.9382} & 0.0040 & 0.8404 & 0.0057 & 0.8532 & 0.0051 & 0.8513 & 0.0046 & 0.8155 & 0.0037 \\
 & 60 & \textbf{0.8715} & 0.0062 & \textbf{0.9386} & 0.0037 & 0.8392 & 0.0047 & 0.8523 & 0.0052 & 0.8506 & 0.0040 & 0.8015 & 0.0033 \\
 & 80 & \textbf{0.8731} & 0.0039 & \textbf{0.9395} & 0.0038 & 0.8398 & 0.0045 & 0.8535 & 0.0033 & 0.8500 & 0.0040 & 0.8021 & 0.0037 \\
 & 100 & \textbf{0.8735} & 0.0053 & \textbf{0.9416} & 0.0041 & 0.8447 & 0.0064 & 0.8562 & 0.0040 & 0.8513 & 0.0032 & 0.8006 & 0.0032 \\\hline
 \multicolumn{2}{r}{\textbf{Average:}} & \multicolumn{1}{r}{\textbf{0.8433}} & \multicolumn{1}{r}{\textbf{0.0080}} & \multicolumn{1}{r}{\textbf{0.8665}} & \multicolumn{1}{r}{\textbf{0.0063}} & \multicolumn{1}{r}{\textbf{0.8313}} & \multicolumn{1}{r}{\textbf{0.0069}} & \multicolumn{1}{r}{\textbf{0.8397}} & \multicolumn{1}{r}{\textbf{0.0075}} & \multicolumn{1}{r}{\textbf{0.8383}} & \multicolumn{1}{r}{\textbf{0.0076}} & \multicolumn{1}{r}{\textbf{0.8111}} & \multicolumn{1}{r}{\textbf{0.0057}} \\
\end{tabular}
}
\end{table}

\begin{table}[ht]
\centering
\caption{Average ranks of balanced accuracies}
\label{tableRanking2}
\renewcommand{\tabcolsep}{1.8mm} 
\renewcommand{\arraystretch}{1.4} 
{\small
\begin{tabular}{|l|rr|rrrr|}
\hline
\multicolumn{1}{|c|}{$K$} & \multicolumn{1}{c}{\textbf{F1}} & \multicolumn{1}{c}{\textbf{F1}} & \multicolumn{1}{c}{BFT} & \multicolumn{1}{c}{HC-CON} & \multicolumn{1}{c}{HC-UWA} & \multicolumn{1}{c|}{FE} \\
 &  & \multicolumn{1}{c}{{\scriptsize  ($\theta$-weighted)}} &  & & & \\
\hline
40       &  \textbf{2.98} &        \textbf{2.37} &           3.94 &           3.33 &           3.43 &           4.95 \\
60       &  \textbf{2.88} &        \textbf{2.45} &           3.97 &           3.29 &           3.40 &           5.02 \\
80       &  \textbf{2.92} &        \textbf{2.39} &           3.97 &           3.25 &           3.38 &           5.09 \\
100      &  \textbf{2.97} &        \textbf{2.41} &           4.00 &           3.20 &           3.36 &           5.06 \\\hline
\multicolumn{1}{r}{\textbf{Avg:}} &  \multicolumn{1}{r}{\textbf{2.94}} & \multicolumn{1}{r}{\textbf{2.40}} & \multicolumn{1}{r}{\textbf{3.97}} & \multicolumn{1}{r}{\textbf{3.27}} & \multicolumn{1}{r}{\textbf{3.39}} & \multicolumn{1}{r}{\textbf{5.03}}
\end{tabular}
}
\end{table}

As opposed to the results shown in Section \ref{sec:resultsAccuracy}, here our formulation with both objective functions outperformed the benchmarks. Moreover, using the $\theta$-weighted function as objective resulted in consistent outperformance over \textbf{F1} and all the benchmarks, with better overall average accuracy, lower dispersion and better ranks, especially for the larger (and more imbalanced) datasets. \textbf{F1} ($\theta$-weighted) outperformed all benchmarks in 22 out of 36 cases. We also note that it had worse performance than the benchmarks in MSK and DRD, which are the most balanced datasets. These results suggest that being able to configure the objective function according to the characteristics of the dataset at hand can be highly beneficial.

\section{Conclusions and future directions}
\label{sec:conclusion}

In this work we proposed an IP approach for the problem of selecting a subset of classifiers in ensemble learning. The objective is to maximise a weighted function of the patterns in the confusion matrix. In order to combine performance and diversity criteria, we also proposed linear constraints to enforce minimum diversity levels. We observed that state-of-the-art solvers can find good solutions in reasonable computational times for the chosen publicly available datasets. The IP approach is, in our view, able to provide a flexible exact algorithm (with regards to both the choice of performance metric and desired diversity levels) which can also be used as a heuristic if short computational time limits are required. This approach has the additional advantage of providing bounds on optimal values.

We compared our formulation to seven well-known benchmarks. We used a stratified 10-fold cross validation procedure and evaluated the effect of enforcing minimum diversity levels and varying the weights assignments of the objective function. In our view the results suggest that our approach is competitive and its flexibility can be beneficial when adapting to datasets with different characteristics. To help future research all the data required to reproduce our results is made available as supplementary material.

As future work we intend to experiment with different criteria and larger datasets. We also intend to research IP techniques and formulation-based heuristics for both finding good solutions quickly and solving the formulation faster. Other lines of research include the study of alternative linear diversity constraints based on different criteria.

\bibliographystyle{plainnat}
\bibliography{references}

\end{document}


\maketitle

In this supplementary material we include three sections. In Section \ref{sec:results}, we provide the full computational results, where we compare our formulation to seven benchmarks instead of four, as reported in the main paper. In Section \ref{sec:classifiers}, we describe all the classifiers used to generate the $A$ and $B$ matrices described in the main paper and used as input in both our formulation and the benchmarks. Section \ref{sec:datasets} provides a link where all test instances can be downloaded. Their format is also explained.

\section{Full computational results}
\label{sec:results}

In Section 5.2 of the main paper, we mention that our proposed formulation is compared to seven other approaches. In the paper, however, we include only four of them due to space constraints. In this section we include the full results with all seven benchmarks. In summary the benchmarks are Full Ensemble (FE), Backfitting (BFT) \citep{friedman1981}, Kappa-Pruning (KP) and four hill climbing methods with different fitness functions: Accuracy (HC-ACC), Complementariness (HC-COM) \citep{martinez2004}, Concurrency  (HC-CON) \citep{banfield2005} and Uncertainty Weighted Accuracy (HC-UWA) \citep{partalas2010}.

\subsection{Accuracy}

The first set of results is the one maximising accuracy, in which we employed the formulation without and with diversity constraints, as explained in the paper. In summary we employ {\bf F1}, without any diversity constraints, {\bf F2}, with diversity constraints on the minimum average PFC value and {\bf F3}, which enforces both minimum average and minimum individual PFC values.

\begin{sidewaystable}
\centering
\caption{Out-of-sample average accuraries and standard deviations, all benchmarks.}
\label{table1}
\renewcommand{\tabcolsep}{1.1mm} 
\renewcommand{\arraystretch}{1.3} 
{\scriptsize
\begin{tabular}{|lr|rr|rr|rr|rr|rr|rr|rr|rr|rr|rr|}
\hline
\multirow{2}{*}{Dataset} & \multirow{2}{*}{$K$} & \multicolumn{2}{c|}{\textbf{F1}} & \multicolumn{2}{c|}{\textbf{F2}} & \multicolumn{2}{c|}{\textbf{F3}} & \multicolumn{2}{c|}{KP} & \multicolumn{2}{c|}{BFT} & \multicolumn{2}{c|}{HC-ACC} & \multicolumn{2}{c|}{HC-CON} & \multicolumn{2}{c|}{HC-COM} & \multicolumn{2}{c|}{HC-UWA} & \multicolumn{2}{c|}{FE} \\
  &  &  \multicolumn{1}{c}{Avg.} &  \multicolumn{1}{c|}{Std.} &  \multicolumn{1}{c}{Avg.} &  \multicolumn{1}{c|}{Std.} &  \multicolumn{1}{c}{Avg.} &  \multicolumn{1}{c|}{Std.} &  \multicolumn{1}{c}{Avg.} &  \multicolumn{1}{c|}{Std.} &  \multicolumn{1}{c}{Avg.} &  \multicolumn{1}{c|}{Std.} &  \multicolumn{1}{c}{Avg.} &  \multicolumn{1}{c|}{Std.} &  \multicolumn{1}{c}{Avg.} &  \multicolumn{1}{c|}{Std.} &  \multicolumn{1}{c}{Avg.} &  \multicolumn{1}{c|}{Std.} &  \multicolumn{1}{c}{Avg.} &  \multicolumn{1}{c|}{Std.} &  \multicolumn{1}{c}{Avg.} &  Std. \\\hline
PRK & 40 & 0.9066 & 0.0145 &  \textbf{0.9134} & 0.0136 & 0.9114 & 0.0130 & 0.8907 & 0.0149 & 0.9041 & 0.0156 & 0.9073 & 0.0150 & 0.9118 & 0.0152 & 0.9113 & 0.0167 & 0.9098 & 0.0183 & 0.9014 & 0.0066 \\
  & 60 & 0.9036 & 0.0124 &  \textbf{0.9089} & 0.0103 & 0.9084 & 0.0113 & 0.8802 & 0.0086 & 0.8999 & 0.0146 & 0.9059 & 0.0183 & 0.9083 & 0.0176 & 0.9089 & 0.0175 & 0.9089 & 0.0195 & 0.8994 & 0.0088 \\
  & 80 & 0.9051 & 0.0144 & 0.9052 & 0.0124 & 0.9056 & 0.0131 & 0.8837 & 0.0126 & 0.9015 & 0.0084 & 0.9068 & 0.0195 & 0.9048 & 0.0167 & 0.9048 & 0.0180 & 0.9129 & 0.0223 & 0.9004 & 0.0091 \\
  & 100 & 0.9036 & 0.0154 & 0.9070 & 0.0202 & 0.9042 & 0.0183 & 0.8803 & 0.0091 & 0.9020 & 0.0080 & 0.9063 & 0.0202 & 0.9089 & 0.0174 & 0.9058 & 0.0178 & 0.9120 & 0.0213 & 0.8959 & 0.0092 \\\hline
BCW & 40 & 0.9664 & 0.0051 &  \textbf{0.9689} & 0.0052 &  \textbf{0.9690} & 0.0054 & 0.9608 & 0.0067 & 0.9652 & 0.0058 & 0.9664 & 0.0022 & 0.9680 & 0.0032 & 0.9669 & 0.0055 & 0.9671 & 0.0050 & 0.9632 & 0.0058 \\
  & 60 & 0.9676 & 0.0043 &  \textbf{0.9701} & 0.0036 &  \textbf{0.9703} & 0.0038 & 0.9618 & 0.0053 & 0.9671 & 0.0056 & 0.9683 & 0.0049 & 0.9696 & 0.0045 & 0.9689 & 0.0057 & 0.9680 & 0.0040 & 0.9629 & 0.0056 \\
  & 80 & 0.9643 & 0.0057 & 0.9673 & 0.0032 & 0.9673 & 0.0032 & 0.9629 & 0.0029 & 0.9660 & 0.0049 & 0.9697 & 0.0060 & 0.9710 & 0.0048 & 0.9696 & 0.0047 & 0.9701 & 0.0049 & 0.9629 & 0.0051 \\
  & 100 & 0.9618 & 0.0052 & 0.9676 & 0.0036 & 0.9676 & 0.0040 & 0.9636 & 0.0039 & 0.9673 & 0.0047 & 0.9687 & 0.0061 & 0.9710 & 0.0052 & 0.9685 & 0.0043 & 0.9692 & 0.0049 & 0.9627 & 0.0056 \\\hline
MSK & 40 & 0.9109 & 0.0110 & 0.9165 & 0.0093 & 0.9127 & 0.0090 & 0.8923 & 0.0088 & 0.9117 & 0.0122 & 0.9201 & 0.0098 & 0.9238 & 0.0092 & 0.9222 & 0.0088 & 0.9192 & 0.0126 & 0.8896 & 0.0100 \\
  & 60 & 0.9119 & 0.0086 & 0.9129 & 0.0092 & 0.9115 & 0.0091 & 0.8881 & 0.0128 & 0.9083 & 0.0136 & 0.9192 & 0.0118 & 0.9230 & 0.0103 & 0.9207 & 0.0072 & 0.9177 & 0.0114 & 0.8869 & 0.0097 \\
  & 80 & 0.9192 & 0.0097 & 0.9208 & 0.0078 & 0.9208 & 0.0076 & 0.8946 & 0.0140 & 0.9163 & 0.0117 & 0.9204 & 0.0123 & 0.9291 & 0.0090 & 0.9194 & 0.0059 & 0.9207 & 0.0096 & 0.8894 & 0.0099 \\
  & 100 & 0.9196 & 0.0066 & 0.9192 & 0.0075 & 0.9188 & 0.0085 & 0.8906 & 0.0122 & 0.9098 & 0.0135 & 0.9211 & 0.0127 & 0.9299 & 0.0081 & 0.9213 & 0.0095 & 0.9215 & 0.0077 & 0.8885 & 0.0096 \\\hline
QSR & 40 &  \textbf{0.8740} & 0.0041 & 0.8731 & 0.0038 &  \textbf{0.8742} & 0.0047 & 0.8734 & 0.0044 & 0.8719 & 0.0046 & 0.8734 & 0.0038 & 0.8734 & 0.0049 & 0.8723 & 0.0040 & 0.8733 & 0.0040 & 0.8706 & 0.0033 \\
  & 60 &  \textbf{0.8755} & 0.0056 &  \textbf{0.8756} & 0.0045 &  \textbf{0.8740} & 0.0055 & 0.8708 & 0.0048 & 0.8710 & 0.0050 & 0.8725 & 0.0056 & 0.8736 & 0.0077 & 0.8720 & 0.0050 & 0.8725 & 0.0051 & 0.8692 & 0.0039 \\
  & 80 & 0.8715 & 0.0047 & 0.8728 & 0.0047 & 0.8740 & 0.0033 & 0.8714 & 0.0038 & 0.8747 & 0.0045 & 0.8739 & 0.0050 & 0.8731 & 0.0062 & 0.8703 & 0.0068 & 0.8747 & 0.0055 & 0.8706 & 0.0025 \\
  & 100 & 0.8722 & 0.0054 &  \textbf{0.8757} & 0.0027 & 0.8732 & 0.0042 & 0.8739 & 0.0053 & 0.8729 & 0.0050 & 0.8737 & 0.0059 & 0.8724 & 0.0040 & 0.8704 & 0.0048 & 0.8738 & 0.0060 & 0.8705 & 0.0028 \\\hline
DRD & 40 & 0.7401 & 0.0060 & 0.7405 & 0.0091 & 0.7389 & 0.0075 & 0.7313 & 0.0085 & 0.7415 & 0.0080 & 0.7448 & 0.0083 & 0.7418 & 0.0077 & 0.7456 & 0.0084 & 0.7467 & 0.0070 & 0.7086 & 0.0080 \\
  & 60 & 0.7458 & 0.0114 & 0.7462 & 0.0096 & 0.7448 & 0.0104 & 0.7368 & 0.0093 & 0.7475 & 0.0100 & 0.7438 & 0.0090 & 0.7480 & 0.0086 & 0.7450 & 0.0078 & 0.7497 & 0.0068 & 0.7110 & 0.0077 \\
  & 80 & 0.7449 & 0.0063 & 0.7412 & 0.0054 & 0.7437 & 0.0068 & 0.7134 & 0.0108 & 0.7464 & 0.0060 & 0.7467 & 0.0080 & 0.7501 & 0.0065 & 0.7470 & 0.0076 & 0.7492 & 0.0070 & 0.7094 & 0.0078 \\
  & 100 & 0.7401 & 0.0083 & 0.7459 & 0.0054 & 0.7442 & 0.0065 & 0.7144 & 0.0139 & 0.7441 & 0.0065 & 0.7473 & 0.0053 & 0.7486 & 0.0079 & 0.7459 & 0.0057 & 0.7516 & 0.0056 & 0.7114 & 0.0076 \\\hline
SPA & 40 & 0.9529 & 0.0014 & 0.9534 & 0.0013 & 0.9534 & 0.0013 & 0.9455 & 0.0011 & 0.9529 & 0.0008 & 0.9531 & 0.0012 & 0.9535 & 0.0012 & 0.9520 & 0.0021 & 0.9523 & 0.0011 & 0.9476 & 0.0010 \\
  & 60 & 0.9533 & 0.0013 & 0.9542 & 0.0015 & 0.9537 & 0.0017 & 0.9447 & 0.0011 & 0.9535 & 0.0014 & 0.9533 & 0.0016 & 0.9547 & 0.0014 & 0.9529 & 0.0014 & 0.9534 & 0.0012 & 0.9469 & 0.0007 \\
  & 80 & 0.9538 & 0.0018 & 0.9536 & 0.0020 & 0.9536 & 0.0014 & 0.9450 & 0.0010 & 0.9532 & 0.0014 & 0.9543 & 0.0018 & 0.9554 & 0.0016 & 0.9529 & 0.0018 & 0.9536 & 0.0018 & 0.9468 & 0.0008 \\
  & 100 & 0.9535 & 0.0018 & 0.9538 & 0.0016 & 0.9537 & 0.0016 & 0.9446 & 0.0011 & 0.9530 & 0.0014 & 0.9538 & 0.0020 & 0.9557 & 0.0012 & 0.9526 & 0.0013 & 0.9537 & 0.0015 & 0.9467 & 0.0006 \\\hline
DEF & 40 & 0.8206 & 0.0005 & 0.8205 & 0.0005 & 0.8204 & 0.0005 & 0.8206 & 0.0005 & 0.8206 & 0.0004 & 0.8208 & 0.0004 & 0.8205 & 0.0005 & 0.8205 & 0.0005 & 0.8211 & 0.0004 & 0.8207 & 0.0004 \\
  & 60 & 0.8205 & 0.0005 & 0.8204 & 0.0005 & 0.8206 & 0.0003 & 0.8209 & 0.0003 & 0.8207 & 0.0003 & 0.8209 & 0.0003 & 0.8204 & 0.0004 & 0.8206 & 0.0004 & 0.8211 & 0.0004 & 0.8205 & 0.0006 \\
  & 80 & 0.8203 & 0.0006 & 0.8206 & 0.0006 & 0.8206 & 0.0008 & 0.8201 & 0.0008 & 0.8203 & 0.0004 & 0.8206 & 0.0004 & 0.8206 & 0.0004 & 0.8206 & 0.0003 & 0.8212 & 0.0003 & 0.8203 & 0.0004 \\
  & 100 & 0.8202 & 0.0004 & 0.8200 & 0.0003 & 0.8205 & 0.0006 & 0.8198 & 0.0010 & 0.8206 & 0.0004 & 0.8206 & 0.0004 & 0.8205 & 0.0004 & 0.8205 & 0.0003 & 0.8213 & 0.0003 & 0.8201 & 0.0004 \\\hline
BMK & 40 &  \textbf{0.9178} & 0.0004 & 0.9171 & 0.0006 & 0.9170 & 0.0006 & 0.9132 & 0.0005 & 0.9166 & 0.0006 & 0.9170 & 0.0005 & 0.9168 & 0.0007 & 0.9169 & 0.0008 & 0.9173 & 0.0005 & 0.9143 & 0.0007 \\
  & 60 &  \textbf{0.9175} & 0.0005 &  \textbf{0.9173} & 0.0006 &  \textbf{0.9173} & 0.0006 & 0.9117 & 0.0004 & 0.9167 & 0.0003 & 0.9172 & 0.0006 & 0.9170 & 0.0007 & 0.9170 & 0.0007 & 0.9173 & 0.0005 & 0.9134 & 0.0007 \\
  & 80 &  \textbf{0.9174} & 0.0007 &  \textbf{0.9174} & 0.0007 & 0.9172 & 0.0005 & 0.9109 & 0.0005 & 0.9165 & 0.0006 & 0.9169 & 0.0006 & 0.9169 & 0.0009 & 0.9170 & 0.0009 & 0.9173 & 0.0006 & 0.9131 & 0.0005 \\
  & 100 &  \textbf{0.9176} & 0.0004 &  \textbf{0.9174} & 0.0005 & 0.9173 & 0.0007 & 0.9101 & 0.0004 & 0.9167 & 0.0004 & 0.9170 & 0.0006 & 0.9168 & 0.0007 & 0.9169 & 0.0008 & 0.9173 & 0.0008 & 0.9124 & 0.0006 \\\hline
APS & 40 & 0.9937 & 0.0002 & 0.9938 & 0.0002 & 0.9938 & 0.0002 & 0.9933 & 0.0002 & 0.9936 & 0.0002 & 0.9938 & 0.0001 & 0.9938 & 0.0001 & 0.9936 & 0.0002 & 0.9938 & 0.0002 & 0.9929 & 0.0001 \\
  & 60 &  \textbf{0.9938} & 0.0002 &  \textbf{0.9938} & 0.0002 &  \textbf{0.9938} & 0.0002 & 0.9929 & 0.0002 & 0.9936 & 0.0002 & 0.9937 & 0.0001 & 0.9937 & 0.0001 & 0.9936 & 0.0002 & 0.9938 & 0.0001 & 0.9925 & 0.0001 \\
  & 80 &  \textbf{0.9939} & 0.0001 & 0.9938 & 0.0001 & 0.9938 & 0.0001 & 0.9925 & 0.0001 & 0.9936 & 0.0002 & 0.9938 & 0.0002 & 0.9938 & 0.0001 & 0.9936 & 0.0002 & 0.9937 & 0.0002 & 0.9925 & 0.0001 \\
  & 100 & 0.9938 & 0.0001 &  \textbf{0.9939} & 0.0001 &  \textbf{0.9939} & 0.0001 & 0.9924 & 0.0001 & 0.9937 & 0.0002 & 0.9938 & 0.0002 & 0.9938 & 0.0001 & 0.9936 & 0.0002 & 0.9938 & 0.0001 & 0.9925 & 0.0001 \\\hline
 \multicolumn{2}{r}{\textbf{Average:}} &  \multicolumn{1}{r}{\textbf{0.8985}} &  \multicolumn{1}{r}{\textbf{0.0049}} &  \multicolumn{1}{r}{\textbf{0.8997}} &  \multicolumn{1}{r}{\textbf{0.0045}} &  \multicolumn{1}{r}{\textbf{0.8993}} &  \multicolumn{1}{r}{\textbf{0.0046}} &  \multicolumn{1}{r}{\textbf{0.8893}} &  \multicolumn{1}{r}{\textbf{0.0051}} &  \multicolumn{1}{r}{\textbf{0.8979}} &  \multicolumn{1}{r}{\textbf{0.0049}} &  \multicolumn{1}{r}{\textbf{0.8999}} &  \multicolumn{1}{r}{\textbf{0.0056}} &  \multicolumn{1}{r}{\textbf{0.9012}} &  \multicolumn{1}{r}{\textbf{0.0052}} &  \multicolumn{1}{r}{\textbf{0.8998}} &  \multicolumn{1}{r}{\textbf{0.0051}} &  \multicolumn{1}{r}{\textbf{0.9009}} &  \multicolumn{1}{r}{\textbf{0.0055}} &  \multicolumn{1}{r}{\textbf{0.8894}} &  \multicolumn{1}{r}{\textbf{0.0041}}

\end{tabular}
}
\end{sidewaystable}

\begin{table}[ht]
\centering
\caption{Average ranks of accuracies, all benchmarks}
\label{tableRanking1}
\renewcommand{\tabcolsep}{1.8mm} 
\renewcommand{\arraystretch}{1.4} 
{\small
\begin{tabular}{|l|rrr|rrrrrrr|}
\hline
$K$ &      \textbf{F1} &      \textbf{F2} &      \textbf{F3} &  KP &  BFT &  HC-ACC &  HC-CON &  HC-COM &  HC-UWA &  FE \\\hline
40       &           5.16 &           5.04 &           5.11 &           6.94 &           5.44 &           4.98 &           5.01 &           5.24 &           4.91 &           7.17 \\
60       &           5.11 &           4.93 &           5.02 &           7.15 &           5.41 &           5.05 &           4.88 &           5.22 &           4.87 &           7.35 \\
80       &           5.09 &           5.04 &           4.99 &           7.55 &           5.44 &           4.97 &           4.66 &           5.25 &           4.73 &           7.29 \\
100      &           5.20 &           4.93 &           4.94 &           7.56 &           5.41 &           4.97 &           4.70 &           5.24 &           4.68 &           7.37 \\\hline
\multicolumn{1}{c}{\textbf{Avg:}} &  \multicolumn{1}{r}{\textbf{5.14}} &  \multicolumn{1}{r}{\textbf{4.98}} &  \multicolumn{1}{r}{\textbf{5.02}} &  \multicolumn{1}{r}{\textbf{7.30}} &  \multicolumn{1}{r}{\textbf{5.42}} &  \multicolumn{1}{r}{\textbf{4.99}} &  \multicolumn{1}{r}{\textbf{4.81}} &  \multicolumn{1}{r}{\textbf{5.24}} &  \multicolumn{1}{r}{\textbf{4.80}} &  \multicolumn{1}{r}{\textbf{7.29}} 
\end{tabular}
}
\end{table}

The full set of results in shown in Table \ref{table1} and the average ranks are shown in Table \ref{tableRanking1}. The results are consistent with the discussion present in the main paper, in which both our formulations and the benchmarks had very similar performance, with hill climbing based methods having better overall performance, although by a relatively small margin. Among the benchmarks KP had the worst performance, perhaps due to being the only method that optimises a diversity measure instead of a performance-based measure or a combination of both. 

The use of integer programming, which attempts to find optimal solutions, may overfit the selected ensemble to the dataset used in the optimisation. To have an overall idea of overfitting, we compare the average accuracy in out-of-sample test data from Table \ref{table1} with the accuracy of the same selected classifiers, but in the in-sample validation data used in the optimisation approach. 

The comparison is shown in Table \ref{tableOverfitting}. In the {\bf Avg.} column, we perform this comparison by calculating the (average out-of-sample accuracy - average in-sample accuracy) / (average in-sample accuracy). As the results are averages of 100 executions, we also include the standard deviation in column \textbf{Std.}. The last row includes the average of all datasets and values of $K$ from its respective column. The more negative the values are, the more the accuracy of the out-of-sample decreased related to the in-sample, which may be a sign of overfitting to the in-sample data. 

\begin{sidewaystable}
\centering
\renewcommand{\tabcolsep}{1.1mm} 
\renewcommand{\arraystretch}{1.3} 
\caption{Relative difference between average in-sample and average out-of-sample accuracies}
\label{tableOverfitting}
{\scriptsize
\begin{tabular}{|lr|rr|rr|rr|rr|rr|rr|rr|rr|rr|}
\hline
\multirow{2}{*}{Dataset} & \multirow{2}{*}{$K$} & \multicolumn{2}{c|}{\textbf{F1}} & \multicolumn{2}{c|}{{\bf F2}} & \multicolumn{2}{c|}{{\bf F3}} & \multicolumn{2}{c|}{KP} & \multicolumn{2}{c|}{BFT} & \multicolumn{2}{c|}{HC-ACC} & \multicolumn{2}{c|}{HC-CON} & \multicolumn{2}{c|}{HC-COM} & \multicolumn{2}{c|}{HC-UWA} \\
 & & \multicolumn{1}{c}{Avg.} & \multicolumn{1}{c|}{Std.} & \multicolumn{1}{c}{Avg.} & \multicolumn{1}{c|}{Std.} & \multicolumn{1}{c}{Avg.} & \multicolumn{1}{c|}{Std.} & \multicolumn{1}{c}{Avg.} & \multicolumn{1}{c|}{Std.} & \multicolumn{1}{c}{Avg.} & \multicolumn{1}{c|}{Std.} & \multicolumn{1}{c}{Avg.} & \multicolumn{1}{c|}{Std.} & \multicolumn{1}{c}{Avg.} & \multicolumn{1}{c|}{Std.} & \multicolumn{1}{c}{Avg.} & \multicolumn{1}{c|}{Std.} & \multicolumn{1}{c}{Avg.} & Std. \\
\hline
PRK & 40 & -0.0583 & 0.0223 & -0.0512 & 0.0165 & -0.0534 & 0.0173 & -0.0186 & 0.0240 & -0.0529 & 0.0257 & -0.0522 & 0.0194 & -0.0491 & 0.0237 & -0.0475 & 0.0214 & -0.0474 & 0.0239 \\
 & 60 & -0.0658 & 0.0179 & -0.0604 & 0.0143 & -0.0609 & 0.0170 & -0.0268 & 0.0265 & -0.0598 & 0.0263 & -0.0586 & 0.0238 & -0.0558 & 0.0213 & -0.0537 & 0.0233 & -0.0520 & 0.0273 \\
 & 80 & -0.0667 & 0.0242 & -0.0666 & 0.0220 & -0.0661 & 0.0214 & -0.0248 & 0.0272 & -0.0611 & 0.0178 & -0.0592 & 0.0249 & -0.0624 & 0.0204 & -0.0598 & 0.0239 & -0.0511 & 0.0287 \\
 & 100 & -0.0699 & 0.0237 & -0.0663 & 0.0283 & -0.0693 & 0.0270 & -0.0233 & 0.0278 & -0.0575 & 0.0166 & -0.0607 & 0.0249 & -0.0584 & 0.0225 & -0.0593 & 0.0252 & -0.0525 & 0.0281 \\\hline
BCW & 40 & -0.0244 & 0.0057 & -0.0219 & 0.0062 & -0.0217 & 0.0065 & -0.0043 & 0.0067 & -0.0190 & 0.0083 & -0.0194 & 0.0050 & -0.0190 & 0.0071 & -0.0193 & 0.0085 & -0.0181 & 0.0089 \\
 & 60 & -0.0247 & 0.0050 & -0.0222 & 0.0051 & -0.0221 & 0.0053 & -0.0021 & 0.0045 & -0.0181 & 0.0094 & -0.0185 & 0.0087 & -0.0181 & 0.0081 & -0.0184 & 0.0097 & -0.0185 & 0.0076 \\
 & 80 & -0.0291 & 0.0068 & -0.0261 & 0.0046 & -0.0261 & 0.0046 & -0.0041 & 0.0051 & -0.0196 & 0.0074 & -0.0179 & 0.0090 & -0.0171 & 0.0082 & -0.0186 & 0.0083 & -0.0169 & 0.0086 \\
 & 100 & -0.0323 & 0.0063 & -0.0264 & 0.0044 & -0.0264 & 0.0048 & -0.0041 & 0.0065 & -0.0188 & 0.0070 & -0.0195 & 0.0080 & -0.0180 & 0.0081 & -0.0207 & 0.0058 & -0.0186 & 0.0085 \\\hline
MSK & 40 & -0.0503 & 0.0149 & -0.0445 & 0.0116 & -0.0484 & 0.0121 & 0.0045 & 0.0212 & -0.0344 & 0.0142 & -0.0338 & 0.0133 & -0.0274 & 0.0138 & -0.0270 & 0.0141 & -0.0268 & 0.0180 \\
 & 60 & -0.0518 & 0.0156 & -0.0508 & 0.0131 & -0.0523 & 0.0142 & -0.0003 & 0.0225 & -0.0384 & 0.0167 & -0.0365 & 0.0165 & -0.0290 & 0.0155 & -0.0302 & 0.0122 & -0.0301 & 0.0171 \\
 & 80 & -0.0522 & 0.0152 & -0.0505 & 0.0131 & -0.0505 & 0.0131 & -0.0027 & 0.0236 & -0.0352 & 0.0184 & -0.0411 & 0.0166 & -0.0305 & 0.0099 & -0.0381 & 0.0108 & -0.0338 & 0.0157 \\
 & 100 & -0.0529 & 0.0108 & -0.0535 & 0.0108 & -0.0539 & 0.0138 & -0.0038 & 0.0217 & -0.0404 & 0.0187 & -0.0410 & 0.0173 & -0.0296 & 0.0100 & -0.0360 & 0.0151 & -0.0342 & 0.0135 \\\hline
QSR & 40 & -0.0381 & 0.0066 & -0.0390 & 0.0074 & -0.0377 & 0.0089 & -0.0037 & 0.0103 & -0.0316 & 0.0108 & -0.0294 & 0.0067 & -0.0253 & 0.0086 & -0.0232 & 0.0102 & -0.0252 & 0.0102 \\
 & 60 & -0.0413 & 0.0088 & -0.0412 & 0.0076 & -0.0429 & 0.0079 & -0.0024 & 0.0090 & -0.0341 & 0.0107 & -0.0335 & 0.0095 & -0.0273 & 0.0097 & -0.0253 & 0.0086 & -0.0288 & 0.0081 \\
 & 80 & -0.0499 & 0.0100 & -0.0485 & 0.0108 & -0.0471 & 0.0091 & -0.0038 & 0.0103 & -0.0332 & 0.0078 & -0.0349 & 0.0067 & -0.0306 & 0.0115 & -0.0296 & 0.0117 & -0.0288 & 0.0099 \\
 & 100 & -0.0519 & 0.0088 & -0.0481 & 0.0071 & -0.0508 & 0.0094 & -0.0001 & 0.0118 & -0.0356 & 0.0076 & -0.0382 & 0.0094 & -0.0334 & 0.0099 & -0.0306 & 0.0100 & -0.0324 & 0.0084 \\\hline
DRD & 40 & -0.0730 & 0.0094 & -0.0719 & 0.0137 & -0.0742 & 0.0131 & -0.0083 & 0.0169 & -0.0516 & 0.0124 & -0.0506 & 0.0128 & -0.0458 & 0.0081 & -0.0347 & 0.0106 & -0.0407 & 0.0079 \\
 & 60 & -0.0786 & 0.0153 & -0.0766 & 0.0122 & -0.0786 & 0.0124 & -0.0058 & 0.0145 & -0.0556 & 0.0130 & -0.0600 & 0.0122 & -0.0431 & 0.0132 & -0.0402 & 0.0086 & -0.0427 & 0.0092 \\
 & 80 & -0.0858 & 0.0118 & -0.0866 & 0.0081 & -0.0807 & 0.0103 & -0.0103 & 0.0144 & -0.0638 & 0.0087 & -0.0618 & 0.0100 & -0.0469 & 0.0075 & -0.0410 & 0.0124 & -0.0488 & 0.0072 \\
 & 100 & -0.0983 & 0.0114 & -0.0844 & 0.0103 & -0.0803 & 0.0116 & -0.0118 & 0.0190 & -0.0714 & 0.0094 & -0.0650 & 0.0069 & -0.0524 & 0.0104 & -0.0455 & 0.0108 & -0.0474 & 0.0105 \\\hline
SPA & 40 & -0.0118 & 0.0030 & -0.0112 & 0.0023 & -0.0112 & 0.0024 & -0.0024 & 0.0019 & -0.0092 & 0.0024 & -0.0093 & 0.0028 & -0.0072 & 0.0021 & -0.0074 & 0.0023 & -0.0076 & 0.0027 \\
 & 60 & -0.0130 & 0.0024 & -0.0121 & 0.0025 & -0.0126 & 0.0032 & -0.0017 & 0.0022 & -0.0104 & 0.0026 & -0.0107 & 0.0018 & -0.0072 & 0.0024 & -0.0075 & 0.0022 & -0.0077 & 0.0022 \\
 & 80 & -0.0136 & 0.0024 & -0.0137 & 0.0032 & -0.0137 & 0.0024 & -0.0016 & 0.0023 & -0.0114 & 0.0015 & -0.0104 & 0.0029 & -0.0070 & 0.0026 & -0.0077 & 0.0027 & -0.0084 & 0.0023 \\
 & 100 & -0.0146 & 0.0025 & -0.0142 & 0.0019 & -0.0143 & 0.0020 & -0.0017 & 0.0020 & -0.0120 & 0.0019 & -0.0114 & 0.0030 & -0.0070 & 0.0023 & -0.0080 & 0.0026 & -0.0088 & 0.0020 \\\hline
DEF & 40 & -0.0046 & 0.0016 & -0.0045 & 0.0012 & -0.0045 & 0.0014 & -0.0000 & 0.0009 & -0.0035 & 0.0014 & -0.0034 & 0.0014 & -0.0020 & 0.0010 & -0.0019 & 0.0012 & -0.0023 & 0.0012 \\
 & 60 & -0.0052 & 0.0016 & -0.0049 & 0.0015 & -0.0045 & 0.0014 & 0.0004 & 0.0011 & -0.0039 & 0.0010 & -0.0037 & 0.0014 & -0.0023 & 0.0012 & -0.0020 & 0.0014 & -0.0026 & 0.0014 \\
 & 80 & -0.0054 & 0.0014 & -0.0047 & 0.0019 & -0.0046 & 0.0012 & 0.0002 & 0.0014 & -0.0047 & 0.0010 & -0.0044 & 0.0012 & -0.0021 & 0.0012 & -0.0020 & 0.0012 & -0.0027 & 0.0013 \\
 & 100 & -0.0057 & 0.0015 & -0.0051 & 0.0016 & -0.0042 & 0.0017 & 0.0004 & 0.0013 & -0.0045 & 0.0013 & -0.0047 & 0.0011 & -0.0024 & 0.0012 & -0.0023 & 0.0011 & -0.0028 & 0.0011 \\\hline
BMK & 40 & -0.0032 & 0.0009 & -0.0035 & 0.0010 & -0.0036 & 0.0012 & -0.0001 & 0.0011 & -0.0032 & 0.0011 & -0.0031 & 0.0010 & -0.0018 & 0.0013 & -0.0016 & 0.0013 & -0.0022 & 0.0011 \\
 & 60 & -0.0037 & 0.0009 & -0.0035 & 0.0010 & -0.0034 & 0.0012 & -0.0000 & 0.0011 & -0.0035 & 0.0010 & -0.0032 & 0.0011 & -0.0016 & 0.0012 & -0.0015 & 0.0013 & -0.0024 & 0.0010 \\
 & 80 & -0.0041 & 0.0013 & -0.0034 & 0.0009 & -0.0034 & 0.0011 & -0.0000 & 0.0010 & -0.0042 & 0.0012 & -0.0039 & 0.0012 & -0.0019 & 0.0013 & -0.0016 & 0.0012 & -0.0026 & 0.0009 \\
 & 100 & -0.0040 & 0.0009 & -0.0032 & 0.0009 & -0.0032 & 0.0013 & 0.0001 & 0.0011 & -0.0043 & 0.0008 & -0.0040 & 0.0012 & -0.0021 & 0.0011 & -0.0019 & 0.0011 & -0.0029 & 0.0009 \\\hline
APS & 40 & -0.0012 & 0.0003 & -0.0011 & 0.0003 & -0.0011 & 0.0002 & -0.0000 & 0.0002 & -0.0007 & 0.0002 & -0.0006 & 0.0002 & -0.0005 & 0.0002 & -0.0005 & 0.0002 & -0.0004 & 0.0002 \\
 & 60 & -0.0012 & 0.0002 & -0.0012 & 0.0002 & -0.0012 & 0.0003 & 0.0000 & 0.0002 & -0.0008 & 0.0002 & -0.0007 & 0.0002 & -0.0006 & 0.0001 & -0.0005 & 0.0002 & -0.0004 & 0.0002 \\
 & 80 & -0.0013 & 0.0002 & -0.0013 & 0.0002 & -0.0013 & 0.0001 & 0.0000 & 0.0002 & -0.0009 & 0.0002 & -0.0008 & 0.0002 & -0.0005 & 0.0002 & -0.0006 & 0.0002 & -0.0006 & 0.0001 \\
 & 100 & -0.0014 & 0.0002 & -0.0013 & 0.0001 & -0.0013 & 0.0002 & -0.0000 & 0.0002 & -0.0009 & 0.0002 & -0.0009 & 0.0002 & -0.0006 & 0.0001 & -0.0006 & 0.0002 & -0.0006 & 0.0001 \\\hline
 \multicolumn{2}{r}{\textbf{Average:}} & \multicolumn{1}{r}{\textbf{-0.0330}} & \multicolumn{1}{r}{\textbf{0.0076}} & \multicolumn{1}{r}{\textbf{-0.0313}} & \multicolumn{1}{r}{\textbf{0.0069}} & \multicolumn{1}{r}{\textbf{-0.0314}} & \multicolumn{1}{r}{\textbf{0.0073}} & \multicolumn{1}{r}{\textbf{-0.0045}} & \multicolumn{1}{r}{\textbf{0.0095}} & \multicolumn{1}{r}{\textbf{-0.0253}} & \multicolumn{1}{r}{\textbf{0.0079}} & \multicolumn{1}{r}{\textbf{-0.0252}} & \multicolumn{1}{r}{\textbf{0.0078}} & \multicolumn{1}{r}{\textbf{-0.0213}} & \multicolumn{1}{r}{\textbf{0.0074}} & \multicolumn{1}{r}{\textbf{-0.0207}} & \multicolumn{1}{r}{\textbf{0.0078}} & \multicolumn{1}{r}{\textbf{-0.0208}} & \multicolumn{1}{r}{\textbf{0.0082}} 
\end{tabular}
}
\end{sidewaystable}
 
From the table we can see that the overfitting problem affected the formulations more strongly than the benchmarks. We consider this as expected behaviour since the formulation is solved by seeking optimality, while the other algorithms offer no optimality guarantees. As such, the in-sample accuracies of the formulations tend to be higher than the benchmarks, but deteriorate more out-of-sample. We notice however that including diversity constraints helped reduce this deterioration, and overall the formulations were still competitive out-of-sample, albeit slightly outperformed by the hill climbing methods. This may indicate that with better overfitting control through further studies on how to employ diversity we may be able to improve the out-of-sample results.

\subsection{Balanced accuracy}

The second set of results uses balanced accuracy, which may be more appropriate for imbalanced datasets, as optimisation metric. We employ two different configurations of our objective function, {\bf F1} and {\bf F1} with the $\theta$-weighted function. As shown in the paper maximising the $\theta$-weighted function is equivalent to maximising balanced accuracy. No diversity constraints are employed in these results. The full set of results is shown in Table \ref{table2} and the average ranks are shown in Table \ref{tableRanking2}. 

\begin{sidewaystable}
\centering
\caption{Balanced Accuracy averages and standard deviations}
\label{table2}
\renewcommand{\tabcolsep}{1.1mm} 
\renewcommand{\arraystretch}{1.3} 
{\scriptsize
\begin{tabular}{|lr|rr|rr|rr|rr|rr|rr|rr|rr|rr|}
\hline
\multirow{2}{*}{Dataset} & \multirow{2}{*}{$K$} & \multicolumn{2}{c|}{\textbf{F1}} & \multicolumn{2}{c|}{\textbf{F1} ($\theta$-weighted)} & \multicolumn{2}{c|}{KP} & \multicolumn{2}{c|}{BFT} & \multicolumn{2}{c|}{HC-ACC} & \multicolumn{2}{c|}{HC-CON} & \multicolumn{2}{c|}{HC-COM} & \multicolumn{2}{c|}{HC-UWA} & \multicolumn{2}{c|}{FE} \\
    &          &              \multicolumn{1}{c}{Avg.} &              \multicolumn{1}{c|}{Std.} &                 \multicolumn{1}{c}{Avg.} &              \multicolumn{1}{c|}{Std.} &              \multicolumn{1}{c}{Avg.} &              \multicolumn{1}{c|}{Std.} &              \multicolumn{1}{c}{Avg.} &              \multicolumn{1}{c|}{Std.} &              \multicolumn{1}{c}{Avg.} &              \multicolumn{1}{c|}{Std.} &              \multicolumn{1}{c}{Avg.} &              \multicolumn{1}{c|}{Std.} &              \multicolumn{1}{c}{Avg.} &              \multicolumn{1}{c|}{Std.} &              \multicolumn{1}{c}{Avg.} &              \multicolumn{1}{c|}{Std.} &              \multicolumn{1}{c}{Avg.} &              Std. \\\hline
PRK & 40 &           0.8478 &           0.0328 &              0.8582 &           0.0289 &           0.8078 &           0.0207 &           0.8523 &           0.0177 &           0.8553 &           0.0273 &           0.8643 &           0.0286 &           0.8626 &           0.0282 &           0.8548 &           0.0290 &           0.8308 &           0.0110 \\
    & 60 &           0.8467 &           0.0242 &              0.8545 &           0.0181 &           0.7760 &           0.0136 &           0.8364 &           0.0242 &           0.8544 &           0.0294 &           0.8611 &           0.0309 &           0.8595 &           0.0301 &           0.8552 &           0.0310 &           0.8237 &           0.0158 \\
    & 80 &           0.8464 &           0.0290 &              0.8514 &           0.0204 &           0.7847 &           0.0196 &           0.8433 &           0.0169 &           0.8575 &           0.0276 &           0.8547 &           0.0244 &           0.8554 &           0.0273 &           0.8615 &           0.0331 &           0.8264 &           0.0146 \\
    & 100 &           0.8487 &           0.0306 &              0.8557 &           0.0266 &           0.7758 &           0.0150 &           0.8416 &           0.0103 &           0.8564 &           0.0285 &           0.8599 &           0.0283 &           0.8521 &           0.0249 &           0.8598 &           0.0320 &           0.8154 &           0.0120 \\\hline
BCW & 40 &  \textbf{0.9645} &           0.0057 &     \textbf{0.9668} &           0.0050 &           0.9543 &           0.0079 &           0.9581 &           0.0066 &           0.9602 &           0.0034 &           0.9624 &           0.0040 &           0.9608 &           0.0063 &           0.9608 &           0.0056 &           0.9566 &           0.0077 \\
    & 60 &  \textbf{0.9666} &           0.0052 &     \textbf{0.9675} &           0.0037 &           0.9556 &           0.0065 &           0.9608 &           0.0064 &           0.9622 &           0.0055 &           0.9638 &           0.0047 &           0.9625 &           0.0059 &           0.9620 &           0.0041 &           0.9564 &           0.0073 \\
    & 80 &           0.9646 &           0.0054 &              0.9651 &           0.0044 &           0.9564 &           0.0039 &           0.9601 &           0.0059 &           0.9643 &           0.0072 &           0.9655 &           0.0053 &           0.9636 &           0.0051 &           0.9647 &           0.0059 &           0.9562 &           0.0069 \\
    & 100 &           0.9625 &           0.0050 &              0.9631 &           0.0056 &           0.9575 &           0.0047 &           0.9616 &           0.0059 &           0.9635 &           0.0070 &           0.9658 &           0.0052 &           0.9634 &           0.0048 &           0.9642 &           0.0055 &           0.9562 &           0.0073 \\\hline
MSK & 40 &           0.9103 &           0.0102 &              0.9104 &           0.0083 &           0.8872 &           0.0103 &           0.9071 &           0.0133 &           0.9165 &           0.0094 &           0.9216 &           0.0090 &           0.9192 &           0.0095 &           0.9167 &           0.0129 &           0.8844 &           0.0108 \\
    & 60 &           0.9108 &           0.0080 &              0.9094 &           0.0084 &           0.8834 &           0.0142 &           0.9047 &           0.0143 &           0.9154 &           0.0119 &           0.9207 &           0.0103 &           0.9180 &           0.0079 &           0.9153 &           0.0124 &           0.8815 &           0.0102 \\
    & 80 &           0.9179 &           0.0100 &              0.9164 &           0.0075 &           0.8899 &           0.0150 &           0.9121 &           0.0124 &           0.9163 &           0.0122 &           0.9259 &           0.0095 &           0.9163 &           0.0063 &           0.9176 &           0.0101 &           0.8841 &           0.0105 \\
    & 100 &           0.9180 &           0.0071 &              0.9162 &           0.0058 &           0.8859 &           0.0131 &           0.9056 &           0.0143 &           0.9172 &           0.0127 &           0.9268 &           0.0083 &           0.9183 &           0.0097 &           0.9184 &           0.0082 &           0.8833 &           0.0102 \\\hline
QSR & 40 &  \textbf{0.8574} &           0.0066 &     \textbf{0.8645} &           0.0049 &           0.8489 &           0.0052 &           0.8496 &           0.0052 &           0.8541 &           0.0056 &           0.8533 &           0.0063 &           0.8506 &           0.0031 &           0.8505 &           0.0044 &           0.8432 &           0.0037 \\
    & 60 &  \textbf{0.8604} &           0.0072 &     \textbf{0.8639} &           0.0052 &           0.8432 &           0.0056 &           0.8467 &           0.0051 &           0.8521 &           0.0065 &           0.8533 &           0.0101 &           0.8503 &           0.0064 &           0.8503 &           0.0057 &           0.8407 &           0.0040 \\
    & 80 &  \textbf{0.8559} &           0.0046 &     \textbf{0.8649} &           0.0038 &           0.8450 &           0.0059 &           0.8515 &           0.0050 &           0.8542 &           0.0057 &           0.8532 &           0.0072 &           0.8484 &           0.0074 &           0.8530 &           0.0059 &           0.8421 &           0.0025 \\
    & 100 &  \textbf{0.8563} &           0.0050 &     \textbf{0.8648} &           0.0052 &           0.8478 &           0.0067 &           0.8477 &           0.0049 &           0.8543 &           0.0045 &           0.8530 &           0.0048 &           0.8475 &           0.0053 &           0.8521 &           0.0063 &           0.8419 &           0.0030 \\\hline
DRD & 40 &           0.7407 &           0.0056 &              0.7434 &           0.0062 &           0.7364 &           0.0087 &           0.7465 &           0.0078 &           0.7480 &           0.0085 &           0.7443 &           0.0080 &           0.7481 &           0.0083 &           0.7499 &           0.0073 &           0.7117 &           0.0079 \\
    & 60 &           0.7463 &           0.0115 &              0.7437 &           0.0088 &           0.7416 &           0.0092 &           0.7520 &           0.0098 &           0.7470 &           0.0089 &           0.7508 &           0.0089 &           0.7473 &           0.0077 &           0.7531 &           0.0068 &           0.7145 &           0.0075 \\
    & 80 &           0.7457 &           0.0064 &              0.7468 &           0.0046 &           0.7176 &           0.0105 &           0.7503 &           0.0057 &           0.7494 &           0.0077 &           0.7525 &           0.0065 &           0.7491 &           0.0078 &           0.7525 &           0.0072 &           0.7126 &           0.0077 \\
    & 100 &           0.7404 &           0.0088 &              0.7466 &           0.0093 &           0.7183 &           0.0133 &           0.7479 &           0.0063 &           0.7501 &           0.0056 &           0.7506 &           0.0075 &           0.7478 &           0.0064 &           0.7551 &           0.0057 &           0.7148 &           0.0075 \\\hline
SPA & 40 &  \textbf{0.9493} &           0.0018 &     \textbf{0.9513} &           0.0023 &           0.9384 &           0.0014 &           0.9478 &           0.0012 &           0.9490 &           0.0011 &           0.9490 &           0.0015 &           0.9474 &           0.0030 &           0.9478 &           0.0015 &           0.9415 &           0.0013 \\
    & 60 &           0.9498 &           0.0016 &     \textbf{0.9514} &           0.0020 &           0.9376 &           0.0013 &           0.9489 &           0.0018 &           0.9494 &           0.0020 &           0.9503 &           0.0015 &           0.9487 &           0.0019 &           0.9491 &           0.0015 &           0.9405 &           0.0008 \\
    & 80 &           0.9504 &           0.0021 &     \textbf{0.9521} &           0.0018 &           0.9378 &           0.0012 &           0.9488 &           0.0017 &           0.9507 &           0.0019 &           0.9512 &           0.0017 &           0.9486 &           0.0022 &           0.9493 &           0.0020 &           0.9403 &           0.0009 \\
    & 100 &           0.9500 &           0.0022 &     \textbf{0.9517} &           0.0025 &           0.9372 &           0.0012 &           0.9485 &           0.0020 &           0.9500 &           0.0020 &           0.9515 &           0.0015 &           0.9482 &           0.0020 &           0.9493 &           0.0018 &           0.9402 &           0.0008 \\\hline
DEF & 40 &  \textbf{0.6633} &           0.0022 &     \textbf{0.6949} &           0.0023 &           0.6511 &           0.0016 &           0.6507 &           0.0011 &           0.6530 &           0.0007 &           0.6543 &           0.0024 &           0.6552 &           0.0017 &           0.6559 &           0.0009 &           0.6507 &           0.0011 \\
    & 60 &  \textbf{0.6643} &           0.0020 &     \textbf{0.6973} &           0.0021 &           0.6510 &           0.0009 &           0.6491 &           0.0018 &           0.6525 &           0.0013 &           0.6541 &           0.0017 &           0.6553 &           0.0012 &           0.6556 &           0.0009 &           0.6490 &           0.0012 \\
    & 80 &  \textbf{0.6662} &           0.0014 &     \textbf{0.6985} &           0.0016 &           0.6490 &           0.0029 &           0.6484 &           0.0020 &           0.6515 &           0.0014 &           0.6553 &           0.0009 &           0.6556 &           0.0010 &           0.6557 &           0.0011 &           0.6484 &           0.0011 \\
    & 100 &  \textbf{0.6661} &           0.0019 &     \textbf{0.6992} &           0.0014 &           0.6468 &           0.0041 &           0.6482 &           0.0018 &           0.6512 &           0.0014 &           0.6542 &           0.0020 &           0.6552 &           0.0010 &           0.6560 &           0.0007 &           0.6473 &           0.0012 \\\hline
BMK & 40 &  \textbf{0.7731} &           0.0055 &     \textbf{0.8607} &           0.0015 &           0.6862 &           0.0027 &           0.7331 &           0.0038 &           0.7423 &           0.0024 &           0.7453 &           0.0027 &           0.7476 &           0.0030 &           0.7462 &           0.0025 &           0.6993 &           0.0039 \\
    & 60 &  \textbf{0.7763} &           0.0049 &     \textbf{0.8662} &           0.0019 &           0.6685 &           0.0033 &           0.7338 &           0.0028 &           0.7434 &           0.0032 &           0.7477 &           0.0029 &           0.7486 &           0.0029 &           0.7469 &           0.0019 &           0.6883 &           0.0038 \\
    & 80 &  \textbf{0.7765} &           0.0055 &     \textbf{0.8684} &           0.0015 &           0.6594 &           0.0037 &           0.7322 &           0.0046 &           0.7406 &           0.0040 &           0.7477 &           0.0028 &           0.7479 &           0.0020 &           0.7469 &           0.0018 &           0.6822 &           0.0036 \\
    & 100 &  \textbf{0.7794} &           0.0053 &     \textbf{0.8694} &           0.0012 &           0.6522 &           0.0033 &           0.7363 &           0.0040 &           0.7417 &           0.0035 &           0.7478 &           0.0029 &           0.7477 &           0.0030 &           0.7479 &           0.0018 &           0.6762 &           0.0031 \\\hline
APS & 40 &  \textbf{0.8690} &           0.0062 &     \textbf{0.9382} &           0.0040 &           0.8296 &           0.0064 &           0.8404 &           0.0057 &           0.8506 &           0.0042 &           0.8532 &           0.0051 &           0.8515 &           0.0044 &           0.8513 &           0.0046 &           0.8155 &           0.0037 \\
    & 60 &  \textbf{0.8715} &           0.0062 &     \textbf{0.9386} &           0.0037 &           0.8150 &           0.0070 &           0.8392 &           0.0047 &           0.8489 &           0.0027 &           0.8523 &           0.0052 &           0.8512 &           0.0045 &           0.8506 &           0.0040 &           0.8015 &           0.0033 \\
    & 80 &  \textbf{0.8731} &           0.0039 &     \textbf{0.9395} &           0.0038 &           0.8009 &           0.0053 &           0.8398 &           0.0045 &           0.8498 &           0.0049 &           0.8535 &           0.0033 &           0.8493 &           0.0051 &           0.8500 &           0.0040 &           0.8021 &           0.0037 \\
    & 100 &  \textbf{0.8735} &           0.0053 &     \textbf{0.9416} &           0.0041 &           0.7997 &           0.0055 &           0.8447 &           0.0064 &           0.8516 &           0.0061 &           0.8562 &           0.0040 &           0.8508 &           0.0031 &           0.8513 &           0.0032 &           0.8006 &           0.0032 \\\hline
 \multicolumn{2}{r}{\textbf{Average:}} &  \multicolumn{1}{r}{\textbf{0.8433}} &  \multicolumn{1}{r}{\textbf{0.0080}} &     \multicolumn{1}{r}{\textbf{0.8665}} &  \multicolumn{1}{r}{\textbf{0.0063}} &  \multicolumn{1}{r}{\textbf{0.8076}} &  \multicolumn{1}{r}{\textbf{0.0073}} &  \multicolumn{1}{r}{\textbf{0.8313}} &  \multicolumn{1}{r}{\textbf{0.0069}} &  \multicolumn{1}{r}{\textbf{0.8368}} &  \multicolumn{1}{r}{\textbf{0.0077}} &  \multicolumn{1}{r}{\textbf{0.8397}} &  \multicolumn{1}{r}{\textbf{0.0075}} &  \multicolumn{1}{r}{\textbf{0.8375}} &  \multicolumn{1}{r}{\textbf{0.0072}} &  \multicolumn{1}{r}{\textbf{0.8383}} &  \multicolumn{1}{r}{\textbf{0.0076}} &  \multicolumn{1}{r}{\textbf{0.8111}} &  \multicolumn{1}{r}{\textbf{0.0057}}

\end{tabular}
}
\end{sidewaystable}

\begin{table}[ht]
\centering
\caption{Average ranks of balanced accuracies, all benchmarks}
\label{tableRanking2}
\renewcommand{\tabcolsep}{1.8mm} 
\renewcommand{\arraystretch}{1.4} 
{\small
\begin{tabular}{|l|rr|rrrrrrr|}
\hline
$K$ &             \textbf{F1} & \textbf{F1} ($\theta$-weighted) &  KP &    BFT &         HC-ACC &         HC-CON &         HC-COM &         HC-UWA &  FE \\\hline
40       &  \textbf{3.94} &        \textbf{3.18} &           6.77 &           5.51 &           4.77 &           4.53 &           4.63 &           4.64 &           7.03 \\
60       &  \textbf{3.78} &        \textbf{3.25} &           6.90 &           5.51 &           4.85 &           4.41 &           4.62 &           4.57 &           7.12 \\
80       &  \textbf{3.80} &        \textbf{3.16} &           7.23 &           5.45 &           4.82 &           4.27 &           4.70 &           4.50 &           7.08 \\
100      &  \textbf{3.91} &        \textbf{3.20} &           7.17 &           5.51 &           4.76 &           4.25 &           4.68 &           4.47 &           7.05 \\\hline
\multicolumn{1}{r}{\textbf{Avg:}} &  \multicolumn{1}{r}{\textbf{3.86}} &        \multicolumn{1}{r}{\textbf{3.20}} &  \multicolumn{1}{r}{\textbf{7.02}} &  \multicolumn{1}{r}{\textbf{5.49}} &  \multicolumn{1}{r}{\textbf{4.80}} &  \multicolumn{1}{r}{\textbf{4.36}} &  \multicolumn{1}{r}{\textbf{4.65}} &  \multicolumn{1}{r}{\textbf{4.54}} &  \multicolumn{1}{r}{\textbf{7.07}} \\
\end{tabular}
}
\end{table}

The discussion on the balanced accuracy results presented in the main paper are also valid here. {\bf F1}, which maximises accuracy, outperformed all benchmarks. The results were even further improved when considering {\bf F1} with the $\theta$-weighted function, which was consistently better, especially for BMK and APS datasets, which are both the largest and the most imbalanced.

\section{Description of the classifiers}
\label{sec:classifiers}

As mentioned in the main text of the paper, we prepared 10 different heterogeneous classifier models, but used up to $K = 100$ different classifiers for creating ensembles. Each of the 10 models was initiated 10 times with different random seeds and parameters. It is worth reminding that, as mentioned in the paper, we used subsets of all classifiers, varying their sizes as in $K = \{40, 60, 80, 100\}$. We set $K$ as multiples of 10 in order to have an equal number of instantiations of each classifier model.

For each model, their initialisation differs in two ways: parameter settings and random seeds required by the algorithms. The 10 classifier models are shown below. We include the following information: Model Package, Package Class (sklearn package models), GitHub, documentation, version, a list of parameters we varied and if normalisation (when applicable).

\begin{enumerate}

\item XGBoost

\begin{itemize}
    \item Package: xgboost
    \item GitHub: \href{https://github.com/dmlc/xgboost}{https://github.com/dmlc/xgboost}
    \item Documentation: \href{https://xgboost.readthedocs.io/en/latest/}{https://xgboost.readthedocs.io/en/latest/}
    \item Used version: 0.80
    \item Used parameters: seed, objective, booster, num\_boost\_round, eta, max\_depth, subsample, colsample\_bytree, colsample\_bylevel, colsample\_bynode, gamma, min\_child\_weight
\end{itemize}

\item LightGBM

\begin{itemize}
    \item Package: lightgbm
    \item GitHub: \href{https://github.com/microsoft/LightGBM}{https://github.com/microsoft/LightGBM}
    \item Documentation: \href{https://lightgbm.readthedocs.io/en/latest/}{https://lightgbm.readthedocs.io/en/latest/}
    \item Used version: 2.2.0
    \item Used parameters: data\_random\_seed, feature\_fraction\_seed, bagging\_seed, objective, boosting\_type, num\_boost\_round, learning\_rate, max\_depth, num\_leaves, feature\_fraction, bagging\_fraction, bagging\_freq, min\_data\_in\_leaf, max\_cat\_to\_onehot, lambda\_l1, lambda\_l2
\end{itemize}

\item Catboost

\begin{itemize}
    \item Package: catboost
    \item GitHub: \href{https://github.com/catboost/catboost}{https://github.com/catboost/catboost}
    \item Documentation: \href{https://catboost.ai/docs/}{https://catboost.ai/docs/}
    \item Used version: 0.11.1
    \item Used parameters: random\_seed, iterations, cat\_features, learning\_rate, l2\_leaf\_reg,  one\_hot\_max\_size, bootstrap\_type, Bayesian, subsample, depth, rsm
\end{itemize}

\item Adaboost

\begin{itemize}
    \item Package: scikit-learn
    \item Class: sklearn.ensemble.AdaBoostClassifier
    \item GitHub: \href{https://github.com/scikit-learn/scikit-learn}{https://github.com/scikit-learn/scikit-learn}
    \item Documentation: \href{https://scikit-learn.org/stable/modules/generated/sklearn.ensemble.AdaBoostClassifier.html}{https://scikit-learn.org/.../sklearn.ensemble.AdaBoostClassifier.html}
    \item Used version: 0.19.2
    \item Used parameters: random\_state, base\_estimator, n\_estimators, learning\_rate
\end{itemize} 

\item Gradient Boosting

\begin{itemize}
    \item Package: scikit-learn
    \item Class: sklearn.ensemble.GradientBoostingClassifier
    \item GitHub: \href{https://github.com/scikit-learn/scikit-learn}{https://github.com/scikit-learn/scikit-learn}
    \item Documentation: \href{https://scikit-learn.org/stable/modules/generated/sklearn.ensemble.GradientBoostingClassifier.html}{https://scikit-learn.org/.../sklearn.ensemble.GradientBoostingClassifier.html}
    \item Used version: 0.19.2
    \item Used parameters: random\_state, n\_estimators, learning\_rate, max\_depth, criterion, subsample, max\_features, min\_samples\_split, min\_samples\_leaf

\end{itemize}

\item Bagging

\begin{itemize}
    \item Package: scikit-learn
    \item Class: sklearn.ensemble.BaggingClassifier
    \item GitHub: \href{https://github.com/scikit-learn/scikit-learn}{https://github.com/scikit-learn/scikit-learn}
    \item Documentation: \href{https://scikit-learn.org/stable/modules/generated/sklearn.ensemble.BaggingClassifier.html}{https://scikit-learn.org/.../sklearn.ensemble.BaggingClassifier.html}
    \item Used version: 0.19.2
    \item Used parameters: random\_state, base\_estimator, n\_estimators, learning\_rate, max\_features, max\_samples
\end{itemize}

\item Random Forest


\begin{itemize}
    \item Package: scikit-learn
    \item Class: sklearn.ensemble.RandomForestClassifier
    \item GitHub: \href{https://github.com/scikit-learn/scikit-learn}{https://github.com/scikit-learn/scikit-learn}
    \item Documentation: \href{https://scikit-learn.org/stable/modules/generated/sklearn.ensemble.RandomForestClassifier.html}{https://scikit-learn.org/.../sklearn.ensemble.RandomForestClassifier.html}
    \item Used version: 0.19.2
    \item Used parameters: random\_state, n\_estimators, criterion, max\_features, max\_depth, min\_samples\_split, min\_samples\_leaf
\end{itemize}

\item Extra Trees

\begin{itemize}
    \item Package: scikit-learn
    \item Class: sklearn.ensemble.ExtraTreesClassifier
    \item GitHub: \href{https://github.com/scikit-learn/scikit-learn}{https://github.com/scikit-learn/scikit-learn}
    \item Documentation: \href{https://scikit-learn.org/stable/modules/generated/sklearn.ensemble.ExtraTreesClassifier.html}{https://scikit-learn.org/.../sklearn.ensemble.ExtraTreesClassifier.html}
    \item Used version: 0.19.2
    \item Used parameters: random\_state, n\_estimators, criterion, max\_features, max\_depth, min\_samples\_split, min\_samples\_leaf
\end{itemize}

\item Logistic Regression

\begin{itemize}
    \item Package: scikit-learn
    \item Class: sklearn.linear\_model.LogisticRegression
    \item GitHub: \href{https://github.com/scikit-learn/scikit-learn}{https://github.com/scikit-learn/scikit-learn}
    \item Documentation: \href{https://scikit-learn.org/stable/modules/generated/sklearn.linear\_model.LogisticRegression.html}{https://scikit-learn.org/.../sklearn.linear\_model.LogisticRegression.html}
    \item Used version: 0.19.2
    \item Used parameters: random\_state, max\_iter, solver, penalty, dual, fit\_intercept, C, tol 
\end{itemize}

\item Multilayer Perceptron

\begin{itemize}
    \item Package: scikit-learn
    \item Class: sklearn.neural\_network.MLPClassifier.html
    \item GitHub: \href{https://github.com/scikit-learn/scikit-learn}{https://github.com/scikit-learn/scikit-learn}
    \item Documentation: \href{https://scikit-learn.org/stable/modules/generated/sklearn.neural\_network.MLPClassifier.html}{https://scikit-learn.org/.../sklearn.neural\_network.MLPClassifier.html}
    \item Used version: 0.19.2
    \item Used parameters: random\_state, max\_iter, hidden\_layer\_sizes, activation, solver, learning\_rate, learning\_rate\_init, momentum
    \item Normalisation: \href{https://scikit-learn.org/stable/modules/generated/sklearn.preprocessing.StandardScaler.html}{sklearn.preprocessing.StandardScaler} with default parameters
\end{itemize}

\end{enumerate}

As mentioned in the paper, we used a stratified 10-Fold procedure to evaluate the proposed method. To generate different folds we varied the random seed required to shuffle the datasets. 

Random number generation is executed twice (classifier initialisation and dataset shuffling). For the shuffling we used random seeds $10, 20, 30, \ldots, 100$. The seeds used in the classifiers depend on the stratified 10-fold seed. For instance, if $K = 100$ and we used seed 10 for the folds, then the seeds for each of the 10 classifiers of each model were 10, 11, \ldots 19.

\section{Availability of the dataset}
\label{sec:datasets}

We have made all the files required to reproduce our experiments available at the following link:
\vspace{0.2cm}

\url{https://drive.google.com/open?id=13tqqyWc-7EtslfaVG_orR4STUEErN_Sw}

\vspace{0.2cm}
\noindent A Google drive account was created for uploading this dataset in order to maintain anonymity. 

In the link, there is one ZIP file for each of the 9 datasets. The zips for the larger datasets are naturally bigger. The files for APS, BMK and DEF are around 1GB in size. Each ZIP file contains 100 CSV files. The results in Tables 4 and 5 in the paper are all based in these CSV files. For each dataset, we have files for the 10 seeds ($10, 20, \ldots, 100$) and for the 10 folds.

Each file contains, in each row, the data points in the validation and test (independent) sets. (the training set is not required to reproduce our experiments). The real classification (0 or 1) of each data point is given, as well as all the classifications from each individual classifier. The individual classifications are given as numbers between 0 and 1, and need to be rounded to generate the $A$ and $B$ matrices. We used standard rounding (0.5 rounds to 1). The CSV files are organised as follows:

\begin{itemize}
    \item The first column, {\bf datatype}, indicates whether that data point is in the validation set (27\% of the full dataset) or the test (independent) set (10\% of the full dataset).
    \item The second column, {\bf real\_class}, contains the real value of that data point (0 or 1).
    \item The third column, {\bf index}, is a reference to the original dataset, indicating the index of that data point in the original file.
    \item Columns 4 to 103 contain the classifications for each individual classifier. There are 100 classifiers in each CSV file. However, if in a specific experiment $K < 100$ and a multiple of 10, we used only the first $K/10$ columns for each classifier model.
    
    For instance, if $K = 40$, we used classifiers \textbf{xgboost\_1, ..., xgboost\_4,  lightgbm\_1, ..., lightgbm\_4, ...} to compose the full ensemble.
\end{itemize}

The original datasets are available at the URI Machine Learning Repository, which can be found at  \url{https://archive.ics.uci.edu/ml/index.php}. The specific URL for each dataset is given below:

\begin{table*}[ht]
\centering\scriptsize
\begin{tabular}{|lll|}
\multicolumn{1}{c}{Id.} & \multicolumn{1}{c}{Dataset name} & \multicolumn{1}{c}{URL} \\
\hline
PRK & Parkinsons & \url{https://archive.ics.uci.edu/ml/datasets/parkinsons}\\
MSK & Musk (Version 1) & \url{https://archive.ics.uci.edu/ml/datasets/Musk+(Version+1)} \\ 
BCW & Breast Cancer Wisconsin & \url{https://archive.ics.uci.edu/ml/datasets/Breast+Cancer+Wisconsin+(Diagnostic)} \\
QSR & QSAR biodegradation & \url{https://archive.ics.uci.edu/ml/datasets/QSAR+biodegradation} \\
DRD & Diabetic Retinopathy Debrecen &  \url{https://archive.ics.uci.edu/ml/datasets/Diabetic+Retinopathy+Debrecen+Data+Set} \\
SPA & Spambase &  \url{https://archive.ics.uci.edu/ml/datasets/Spambase} \\
DEF & Default of credit card clients & \url{https://archive.ics.uci.edu/ml/datasets/default+of+credit+card+clients} \\
BMK & Bank Marketing & \url{https://archive.ics.uci.edu/ml/datasets/Bank+Marketing} \\
APS & APS Failure at Scania Trucks & \url{https://archive.ics.uci.edu/ml/datasets/APS+Failure+at+Scania+Trucks}\\
\hline
\end{tabular}
\end{table*}

\bibliographystyle{plainnat}
\bibliography{references}